\documentclass[lettersize,journal]{IEEEtran}
\usepackage{amsmath,amsfonts}
\usepackage{algorithmic}
\usepackage{algorithm}
\usepackage{array}
\usepackage{textcomp}
\usepackage{stfloats}
\usepackage{url}
\usepackage{verbatim}
\usepackage{graphicx}
\usepackage{subfig}
\usepackage{cite}
\usepackage{xcolor}
\usepackage{footnote}
\usepackage{threeparttable}
\usepackage{booktabs}
\usepackage{color}
\usepackage{setspace}
\usepackage{epsfig}
\usepackage{epstopdf}
\usepackage{mathrsfs}
\usepackage{accents}
\usepackage{multirow} 
\usepackage{float}    
\usepackage[colorlinks=blue,linkcolor=blue,citecolor=blue]{hyperref}%

\usepackage{ntheorem}
\newtheorem{thm}{Theorem}
\newtheorem{cor}{Corollary}
\theoremstyle{empty}
\newtheorem{reproof}{Proof}

\usepackage{booktabs}
\usepackage{lipsum}
\usepackage{xcolor}
\usepackage{soul}
\sethlcolor{black}
\makeatletter
\newif\if@blind
\@blindtrue 
\if@blind \sethlcolor{black}\else
   
\fi

\begin{document}

\title{ICODE: Modeling Dynamical Systems with Extrinsic Input Information}

    \author{Zhaoyi Li{$^\ast$}, Wenjie Mei{$^\ast$}, \IEEEmembership{Member, IEEE}, Ke Yu, Yang Bai, \IEEEmembership{Member, IEEE}, and Shihua Li, \IEEEmembership{Fellow, IEEE} \thanks{$\ast$ Equal contribution. }
    \thanks{Z. Li, W. Mei, K. Yu, and S. Li are with the School of Automation and the Key Laboratory of MCCSE of the Ministry of Education, Southeast University, Nanjing 210096, China. Y. Bai is with the Graduate School of Advanced Science and Engineering, Hiroshima University, Higashihiroshima 739-0046, Japan. (Corresponding author: Wenjie Mei, email: \href{mailto:wenjie.mei@seu.edu.cn}{\textit{wenjie.mei@seu.edu.cn}}) }}

\markboth{Journal of \LaTeX\ Class Files,~Vol.~14, No.~8, August~2021}%
{Shell \MakeLowercase{\textit{et al.}}: A Sample Article Using IEEEtran.cls for IEEE Journals}


\maketitle

\begin{abstract}
Learning models of dynamical systems with external inputs, which may be, for example, nonsmooth or piecewise, is crucial for studying complex phenomena and predicting future state evolution, which is essential for applications such as safety guarantees and decision-making. In this work, we introduce \emph{Input Concomitant Neural ODEs (ICODEs)}, which incorporate precise real-time input information into the learning process of the models, rather than treating the inputs as hidden parameters to be learned. The sufficient conditions to ensure the model's contraction property are provided to guarantee that system trajectories of the trained model converge to a fixed point, regardless of initial conditions across different training processes. We validate our method through experiments on several representative real dynamics: Single-link robot, DC-to-DC converter, motion dynamics of a rigid body, Rabinovich-Fabrikant equation, Glycolytic-glycogenolytic pathway model, and heat conduction equation. The experimental results demonstrate that our proposed ICODEs efficiently learn the ground truth systems, achieving superior prediction performance under both typical and atypical inputs. This work offers a valuable class of neural ODE models for understanding physical systems with explicit external input information, with potentially promising applications in fields such as physics and robotics.   Our code is available online at \href{https://github.com/EEE-ai59/ICODE.git}{https://github.com/EEE-ai59/ICODE.git}.
\end{abstract}

\def\abstractname{Note to Practitioners} 
\begin{abstract}
Understanding and accurately predicting the behavior of complex dynamical systems is essential for a wide range of practical applications, from robotics and control systems to energy management and biology. Many real-world systems involve external inputs that can be irregular, nonsmooth, or change abruptly, which complicates the modeling process. The method introduced in this paper, Input Concomitant Neural ODEs (ICODEs), offers a new way to integrate these inputs directly into the system model, rather than treating them as unknown or hidden factors. This approach ensures that the model can better reflect real-world conditions, where inputs play a critical role in system behavior. One of the key advantages of ICODEs is their ability to ensure that the system stabilizes to a predictable state, even when trained on different data or starting conditions. This property is particularly useful in safety-critical applications where consistent and reliable system behavior is essential, such as in robotics or in power electronics like DC-to-DC converters. By learning dynamic systems more accurately, ICODEs provide practitioners with a powerful tool for developing systems that can better handle varying inputs while capturing system characteristics. Whether for designing safer robots, optimizing industrial systems, or improving energy efficiency, ICODEs offer promising potential for real-world applications requiring precise modeling of input-driven dynamics.
\end{abstract}


\section{Introduction}
\IEEEPARstart{A}{ccurately} modeling dynamical systems, especially those influenced by external inputs, is fundamental to understanding complex phenomena and predicting their future behavior. Such models are not only significant for advancing theoretical knowledge but are also vital for practical applications that require reliable decision-making \cite{roxin2008neurobiological,brehmer1992dynamic} and safety guarantees \cite{gurriet2020scalable,sidrane2022overt}. Existing related approaches (see, \emph{e.g.}, neural ODEs (NODEs) \cite{chen2018neural} and their variants \cite{dupont2019augmented,kang2021stable,white2024stabilized}) to modeling these systems often ignore the environmental change (input signals and others) or involve treating external inputs as parameters to be learned, which can lead to inaccuracies and inefficiencies, particularly in scenarios where real-time input is important but capricious.  The ability to model these systems with precision directly impacts fields such as:
robotics\cite{10477253,chen2023robust}, economics\cite{yang2023neural}, climate science\cite{irrgang2021towards}, and control engineering\cite{9257478}, where the dynamic interaction between the system and its environment has to be accurately captured. Moreover, the integration of external inputs into dynamical models allows for more realistic modelings, enabling better-informed predictions and the development of robust algorithms capable of adapting to variations.


To address these issues, this paper introduces \emph{ICODEs}, a novel framework that directly incorporates real-time external input data into the learning process of dynamical system models. Unlike conventional learning methods that estimate external inputs as part of the model parameters, ICODEs leverage precise input information during training, leading to more accurate modeling. A key feature of the ICODE framework is its ability to ensure the contraction property of the trained models, despite its possibly high volume of number of subnetworks. This property guarantees that, regardless of the initial conditions, the trajectories of the dynamical system will converge to a fixed point, thereby providing consistency and reliability across different training processes. This is particularly important in applications where stability and predictability are significant. The effectiveness of ICODEs is demonstrated through experiments on several distinct real systems, including single-link robot,  DC-to-DC converter, motion dynamics of a rigid body, Rabinovich-Fabrikant equation, glycolytic-glycogenolytic pathway model, and heat conduction equation. The experimental results show that ICODEs not only achieve superior prediction accuracy for typical inputs but also maintain high performance under atypical input conditions, highlighting their versatility and robustness.

Compared to existing NODE and its variants, \textbf{this work offers several main contributions}: 1) A scalable and flexible NODE framework is proposed for learning dynamical systems with accessible external input information. 2) Theoretical guarantees are provided for the contraction property of ICODEs, ensuring that system trajectories converge to a fixed point regardless of initial conditions. 3) A coupling term between the state and external input is introduced, reflecting the commonly observed interconnection in real physical systems.
4) ICODEs can effectively manage scenarios where external inputs are either typical or atypical, a challenge that previous methods struggled to address.


\subsection*{Related Work} 
\textbf{Neural CDEs and ANODEs vs ICODEs}:   Neural CDEs  \cite{kidger2020neural} (for brevity, we refer to them as "CDEs" throughout the remainder of this paper) are also designed to incorporate incoming information, while Augmented Neural ODEs (ANODEs) \cite{dupont2019augmented} are effective in scenarios where state evolution is less critical. 
A brief introduction to these models and their distinctions from ICODE can be found in appendix A.
However, many real-world models involve complex external environments (such as atypical, for example, piecewise inputs) and couplings of parameters that are challenging to address accurately with existing NODE and its variants, including CDEs and ANODEs. Our experiments demonstrate that ICODEs are better suited for capturing the natural dynamics of systems with complex structural and environmental characteristics. They can be endowed with a significant contraction property despite their intricate structure.

\textbf{Contraction and Stability Property}:
Stability analyses have been performed on NODE variants, such as SODEF \cite{kang2021stable}, SNDEs \cite{white2024stabilized}, LyaNet \cite{rodriguez2022lyanet}, and Stable Neural Flows\cite{massaroli2020stable}. Contraction studies on NODEs from various perspectives are discussed in works such as contractive Hamiltonian NODEs \cite{zakwan2022robust}  and learning stabilizable dynamical systems \cite{singh2018learning}. In addition to contraction analysis, our study emphasizes the scalability aspects related to both the number of hidden layers and the width of neural networks. Additionally, our experiments are primarily concerned with learning and predicting the dynamic behaviors of physical models rather than video processing and image classification \cite{park2021vid,llorente2021deep}.

\textbf{Physical Information Based Learning}: In addition to the method used by ICODEs for incorporating external input (a common presence of physical information), various other models have been developed to ensure that the resulting modeling remains physically plausible in both discrete and continuous time. For example, \cite{bihlo2022physics} employs Physics-Informed Neural Networks (PINNs), which, unlike NODEs and their variants that primarily focus on implicitly learning dynamical systems, explicitly integrate physical laws into the loss function. These approaches have broad applications, for example, PINNs can be used in deep learning frameworks to solve forward and inverse problems involving nonlinear partial differential equations \cite{raissi2019physics}. Additionally, \cite{shi2024towards} and \cite{o2022stochastic} utilize enhanced NODE to learn complex behaviors in physical models. Meanwhile, \cite{greydanus2019hamiltonian} and \cite{cranmer2020lagrangian} introduce Hamiltonian and Lagrangian neural networks, respectively, to learn physically consistent dynamics.


\section{Neural ODEs with External Input}

Let $X$ be a smooth $n$-dimensional manifold, and $U \subset \mathbb{R}^m$ with $m \leq n$. Consider a general dynamical system in the form
\begin{equation} \label{eq:main_sys}
    \dot{x} = f(x,u),
\end{equation}
for representing Neural ODEs (NODEs) with external input, where $x \in X$ is the state, the input $u \in U$, and the vector field $f \colon  X \times U \to X$, \; $(x,u) \to f(x,u)$. More specifically, in this work, we introduce a class of NODEs that are affine in the inputs $u$, where $u=\begin{bmatrix} u_1 & \dots & u_m \end{bmatrix}^\top$, presented as follows. 
\begin{equation} \label{eq:CAS}
    \dot{x} = f_0(x) + \sum_{j=1}^m g_j(x)u_j = f_0(x)+g(x)u,
\end{equation}
where the functions $f_0 \colon X \to \mathbb{R}^n$ and $\begin{bmatrix} g_1 & \dots & g_m  \end{bmatrix} = g \colon X \to \mathbb{R}^{n \times m}$. Here, the cumulative term $\sum_{i=1}^m g_i(x)u_i$ can be regarded as linear combinations of vector fields $g_i$ on $X$ and they constitute a \emph{distribution}, \emph{i.e.}, $\Delta(x) = \text{span} \{g_1(x),\dots,g_m(x)\}$. 


For the practical implementations of the NODE variant~\eqref{eq:CAS}, we introduce  \emph{Input Concomitant Neural ODEs (ICODEs)} with multiple inner layers:

\begin{figure*}[t]
    \centering
\includegraphics[width=0.95\linewidth]{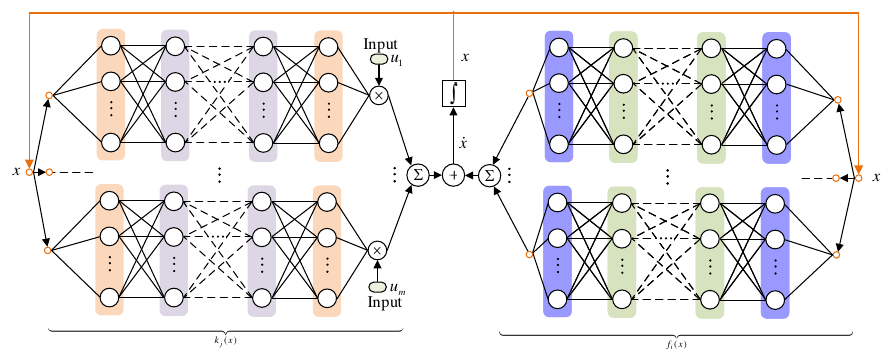}
\caption{The schematic of ICODEs. Using the ODE integration methods, it illustrates how the state $x$ evolves through the external input $u$ and the neural networks (NNs) $k_1, \dots, k_m$, and the NN consisting of the subnetworks $f_1(\cdot)$, ..., $f_M(\cdot)$.}
\label{Fig: flowChart}
\end{figure*}

\begin{align} 
    \dot{x} = & \sum_{i=1}^{M} W_{i0} h_0 (W_{i1} h_1 (\cdots h_{\alpha -1 } (W_{i \alpha}x)))) \nonumber \\
     & + \sum_{j=1}^{m} P_{j0} s_0 (P_{j1} s_1 (\cdots s_{\beta -1 } (P_{j\beta }x))))  u_j, 
\end{align}
where $h_0,\dots,h_{\alpha -1 }, s_0, s_{\beta -1 }$ are the activation functions and $W_{i0}, \dots, W_{i \alpha}, P_{j0}, \dots, P_{j\beta }$ are the weight matrices with appropriate dimensions. For brevity, we let 
\begin{align*} 
     & f_i(x) := W_{i0} h_0 (W_{i1} h_1 (\cdots h_{\alpha -1 } (W_{i \alpha}x)))),  \nonumber \\
     & k_j(x) :=  P_{j0} s_0 (P_{j1} s_1 (\cdots s_{\beta -1 } (P_{j\beta }x)))) 
\end{align*}
for all $x$. Then, the form of~\eqref{eq:CAS} can be transformed into 
\begin{equation} \label{eq:used_NDOE}
\dot{x} = \sum_{i=1}^{M} f_i(x) + \sum_{j=1}^{m} k_j(x) u_j. 
\end{equation}

For neural networks, including ICODEs~\eqref{eq:used_NDOE}, contraction is a critical property that demonstrates the models' learning capability. It refers to the convergence of the models to the theoretical optimal solution throughout the learning process, despite variations in the initial conditions of the training processes. In the following, we show the sufficient conditions of contraction property for~\eqref{eq:used_NDOE}.

\begin{thm} \label{thm:contraction}
The NODEs~\eqref{eq:used_NDOE} are contracting if there exists a uniformly positive definite metric $M(x) = L(x)^\top L(x) $, where $L(x) \in \mathbb{R}^{n \times n}$ is a smooth and invertible transformation, such that for the transformed system of $\delta y = L(x) \delta x$, we have 
\begin{equation} \label{thm:condi}
     \frac{R+R^\top}{2} \leq -cI_{n}
\end{equation}
for some constants $c>0$, where $I_n$ denotes the $n \times n$ identity matrix,
\[
R := L(x)  \Big( \sum_i \frac{\partial f_i }{\partial x} + \sum_j \frac{\partial k_j}{\partial x} u_j \Big) L(x)^{-1} + \dot{L}(x) L(x)^{-1}, 
\]
and the vector $\delta x$ denotes the state error between two neighboring trajectories. 
\end{thm}

\begin{reproof}[Proof of Theorem~1] 
   The proof methodology can be found in \cite{lohmiller1998contraction}. First, differentiating both sides of the transformation $\delta y = L(x) \delta x$, one can obtain 
  \begin{align} 
   \delta \dot{y} & = L \delta \dot{x} + \dot{L} \delta x  \nonumber \\
   & = L J(x,u) (\delta x)  + \dot{L} \delta x, \nonumber 
   \end{align} 
where $J(x,u) := \sum_i \frac{\partial f_i }{\partial x} + \sum_j \frac{\partial k_j}{\partial x} u_j $. We also define $
   R := L J(x,u) L^{-1} + \dot{L} L^{-1}. 
   $
   Then, one has $\delta \dot{y} = R \delta y$. Therefore, we derive
   \[
   \frac{d}{dt} \left(\delta y^\top \delta y \right) = 2 \delta y^\top \delta \dot{y} = 2 \delta y^\top R \delta y.  
   \]
 By the condition~\eqref{thm:condi} in Theorem~\ref{thm:contraction}, we can substantiate that $\delta y$ exponentially converges to $0$.  Since $M$ is uniformly positive definite, the exponential convergence of $\delta y$ to $0$ also implies exponential convergence of $\delta x$ to $0$. This completes the proof. 
\end{reproof}

Furthermore, under the relation $\delta \dot{x} = J(x,u) \delta x $ (recall that $J(x,u) := \sum_i \frac{\partial f_i }{\partial x} + \sum_j \frac{\partial k_j}{\partial x} u_j $), one can directly formulate the following corollary, which displays more restrictive contraction conditions. 

\begin{cor} \label{cor:contraction}
   The NODEs~\eqref{eq:used_NDOE} are contracting if the inequality 
   \[
   \lambda_{\max}\left( \frac{J(x,u) + J(x,u)^\top }{2} \right) \leq -c
   \]
holds true for some $c>0$ and all $x,u$, where $\lambda_{\max}$ denotes the maximum eigenvalue of a matrix. 
\end{cor}

\begin{reproof}[Proof of Corollary~1] 
Under the condition in Corollary~\ref{cor:contraction}, it can be shown that 
\begin{equation*}
\| \delta x(t) \|  \leq  e^{-ct} \| \delta x(0) \|.
\end{equation*}
Thus, we have $\lim_{t \to +\infty} \delta x(t) =0$, demonstrating the contraction property of~\eqref{eq:used_NDOE}. 
\end{reproof}

Our objective is to examine the proposed ICODEs' ability to understand and learn complex physical systems in the context of given external inputs, as in numerous important real-world dynamics, there are accessible real-time input signals that are coupled with the state, motivating the introduction of our models: ICODEs that are "affine" in the input.  


\section{Experimental Results on Learning and Prediction of Physical Systems}\label{experiment}
In this section, we experimentally analyze the performance of ICODEs, compared to CDE \cite{kidger2020neural}, NODEs \cite{chen2018neural}, and ANODEs \cite{dupont2019augmented}. We select several important physical systems in various fields, to test the generality of the proposed ICODEs. We trained the models on time series data and used them to predict the evolution of the future state with known input. The models' training and prediction performance was evaluated.
All experiments were conducted with an NVIDIA GeForce RTX 3080ti GPU under Pytorch 1.11.0, Python 3.8 (Ubuntu 20.04), and 
Cuda  11.3. The detailed description of the hyperparameter setting can be found in appendix C. 

\subsection{Single-link Robot}
In this task, we consider a single-link robot, also known as a one-degree-of-freedom (1-DOF) robotic arm, taking the simple form of a robotic manipulator. It consists of a single rigid arm, which can rotate around a fixed pivot point. Its corresponding dynamics can be represented by
\begin{equation}\label{key7.23-1}
 M \ddot{q}+\frac{1}{2}mgL\sin q =u, \quad 
 y=q,
\end{equation}
where $M$ denotes the moment of inertia, $m$ and $L$ are the mass and length separately, $g$ is the gravity coefficient, $q$ is the angle of the link, and $u$ is the control torque. Suppose that the control torque is pre-designed, and one has to predict the state of the object during system operation. 

\begin{figure*}[t]
\centering
\subfloat[]{
	\label{Fig: slrobot Fig.sub.1}	\includegraphics[width=0.33\textwidth]{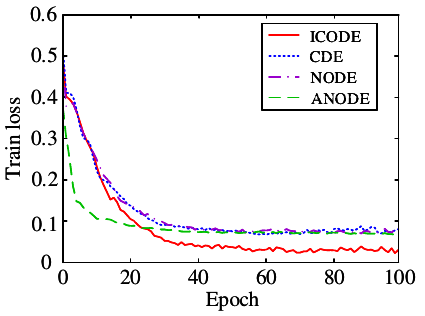}} 
\subfloat[]{
	\label{Fig: slrobot Fig.sub.2}	\includegraphics[width=0.338\textwidth]{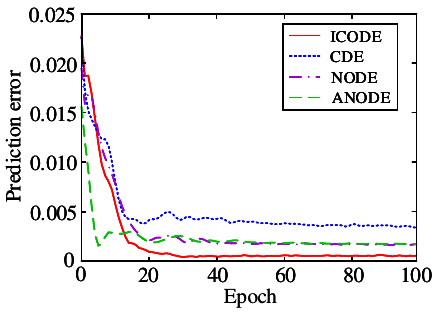}}
\subfloat[]{
	\label{Fig: slrobot Fig.sub.3}
	\includegraphics[width=0.322\textwidth]{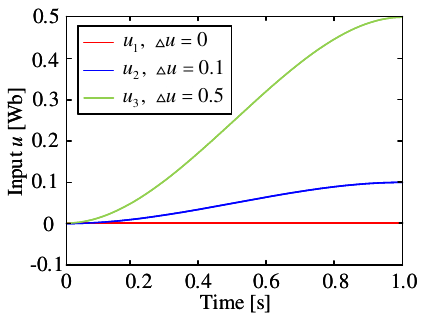}}
\caption{Comparison of training and prediction errors at each time step on the test set for different models on the task of the single-link robot. (a) Training loss in MSE under $u_3$. (b) Prediction loss in MSE under $u_3$. (c) Inputs $u$.}
\label{Fig: slrobot}
\end{figure*}

Let $x_1=q$ and $x_2=\dot{q}$. Then, one can derive the following state space equation
\[
 \dot{x}_1 = x_2, \quad \dot{x}_2 = \frac{1}{M}u - \frac{1}{2M}mgL\sin{x}_1.
\]
We trained on 10 trajectories with initial conditions $x_i(0)$ uniformly sampled between $-1$ and $1$. Each trajectory spans a time period from $0s$ to $1s$ with a constant step:  $\Delta t=0.01$. The parameters were set as follows: $M=1, m=2, L=0.5, g=9.8, q=3.5 \times 10^{-4}$. The smooth control torques $u$ start from $0Wb$, with varying shapes depicted in Fig. \ref{Fig: slrobot Fig.sub.3}. 

The metric results of experiments are summarized in Table \ref{tab: slr}. When $\Delta u=0$ (see the control trajectory in Fig. \ref{Fig: slrobot Fig.sub.3}), indicating the absence of external input, all methods perform comparably well. However, in the second scenario, where $\Delta u=0.1$, ICODE begins to show distinct advantages, outperforming the other methods across all metrics, including \textbf{Mean Squared Error (MSE)}, \textbf{Mean Absolute Error (MAE)}, and \textbf{Coefficient of Determination ($R^2$)}. As $\Delta u$ increases to 0.5, the performance difference widens further, particularly in the $R^2$ score. Our method achieves an $R^2$ of $0.91$, indicating it can explain the majority of the variance in the prediction series, whereas the other competing models perform significantly worse in this regard. The negative $R^2$ scores mean that the other three methods even fail to predict the general trend of the trajectory.

Figs. \ref{Fig: slrobot Fig.sub.1}, \ref{Fig: slrobot Fig.sub.2} show the loss curves of the selected comparative models, under the third input condition ($\Delta u=0.5$). Further statistical details are in Section~\ref{sec:statistical}. Our method performs the best and CDE behaves the worst in the setting. This mainly results from the control torque not affecting the system through its derivatives, thus, the derivatives become an interference that affects the performance of CDE.

\begin{figure}[t]
\centering	
\includegraphics[width=0.5\textwidth]{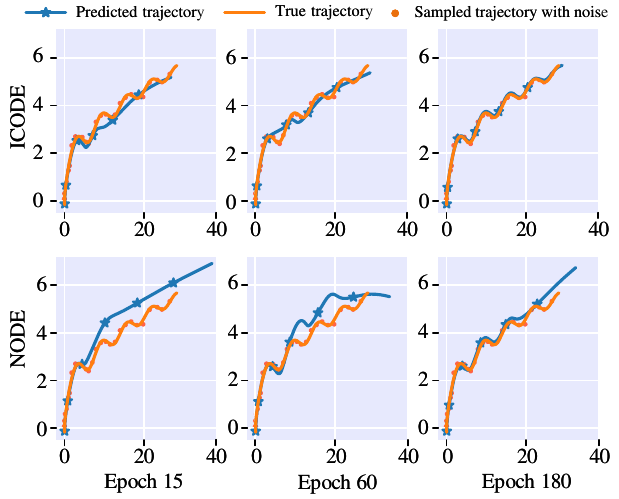}
    \caption{System trajectories of the single-link robot on phase plane (horizontal coordinate: $q$, and vertical coordinate: $\dot{q}$).}
\label{Fig: slrobot-tra}
\end{figure}

Moreover, we set the extended time interval from $0s$ to $10s$ and utilized $1000$ time points for model training. We randomly selected an initial state and plotted the trajectories generated by ICODE and NODE. The resulting state trajectories on the phase plane are presented in Fig. \ref{Fig: slrobot-tra}. The value of input $u$ was chosen to vary between $0Wb$  and $0.5Wb$ at three specific time instants: $0.1s$, $0.4s$, and $0.8s$, remaining constant for the remainder of the time. It is clear that ICODE achieves a faster convergence to the ground truth than NODE. After $15$ epochs, ICODE learns the trend of the original trajectory, while NODE shows noticeable deviations from the true trajectory. When the number of epochs is increased to $180$, ICODE demonstrates a superior ability to accurately fit the true trajectory, whereas NODE's performance is comparatively less effective, particularly in capturing finer details at certain critical points in the phase plane. It is important to note that in this task, we selected a relatively simple input $u$ that remains unchanged within the time interval from $0.8s$ to $10s$. If the shape of $u$ was more complex, the performance gap between ICODE and NODE is expected to increase further.

\begin{table*}
\small
\caption{Performance comparison under various amounts of inputs. } 
\centering
\renewcommand{\arraystretch}{1.2}
\setlength{\tabcolsep}{2.5mm}{
\begin{tabular}{ccccccccc|cccc|ccc}
\toprule
\cmidrule[0.8pt]{1-12} 
& \multicolumn{2}{c|}{}  
 & \multicolumn{9}{c}{Single-link Robot}            \\  \cmidrule{4-12}  
\multicolumn{3}{c|}{Model}                      & \multicolumn{3}{c|}{$\Delta u=0$}              & \multicolumn{3}{c|}{$\Delta u =0.1$}   & \multicolumn{3}{c}{$\Delta u=0.5$}                                   \\ 
\multicolumn{3}{c|}{}                  & RMSE     & MAE   & \multicolumn{1}{c|}{$R^2$} & RMSE     & MAE   & \multicolumn{1}{c|}{$R^2$}        & RMSE     & MAE   & \multicolumn{1}{c}{$R^2$}     \\ 
\midrule
\multicolumn{3}{c|}{ICODE (Ours)}          & 0.027       & 0.065                   & \multicolumn{1}{c|}{0.80}              & \multicolumn{1}{c}{\textbf{0.019}} & \textbf{0.047} &  \multicolumn{1}{c|}{\textbf{0.88}}    & \textbf{0.020} &      \textbf{0.046} &  \textbf{0.91}               \\
\multicolumn{3}{c|}{CDE}             & 0.024     & 0.06                   & \multicolumn{1}{c|}{ 0.77}                & \multicolumn{1}{c}{0.029}  &  0.072 & \multicolumn{1}{c|}{0.33}     & 0.067 & 0.16 &     -11.32          
\\
\multicolumn{3}{c|}{NODE}             &  0.028     & 0.067                   & \multicolumn{1}{c|}{0.80}                & \multicolumn{1}{c}{0.024}  &  0.060 & \multicolumn{1}{c|}{0.62}              & 0.041 &  0.10  &  -9.13     
\\
\multicolumn{3}{c|}{ANODE}             & 0.028    & 0.065                    & \multicolumn{1}{c|}{0.84 }             & \multicolumn{1}{c}{0.026}  &  0.061 & \multicolumn{1}{c|}{0.61}  & 0.044 & 0.10 & -7.26 
\\                       
\bottomrule
\end{tabular}}
\label{tab: slr}
\end{table*}

\subsection{DC-to-DC Converter}
In this experiment, we consider an idealized DC-to-DC converter. Its mathematical expression can be formulated as 
\begin{equation}\label{key7.23-3}
\begin{aligned}
	C_1 \dot{v}_1 & = (1-u)i_3, \\
 C_2 \dot{v}_2 & = ui_3, \\
 L_e \dot{i}_3 & = -(1-u)v_1-uv_2.
 \end{aligned}
\end{equation}
Here, $v_1, v_2$ are the voltages across capacitors $C_1, C_2$, respectively, $i_3$ is the state current across an indicator $L_e$, and $u$ is the control input and its values varying from $0Wb$ to $1Wb$.

We collected 10 trajectories integrated over $1s$, comprising a total of $75$ time points. The initial $50$ points were utilized for training, while the remaining $25$ points were reserved for prediction. The system parameters were configured as $C_1=0.1$, $C_2=0.2$, and $L_e=0.5$.

We examined the learning and prediction ability of ICODEs in the following two scenarios. 

\begin{figure*}
\centering
\subfloat[]{
    \label{Fig: DCDC Fig.sub.1}
    \includegraphics[width=0.33\textwidth]{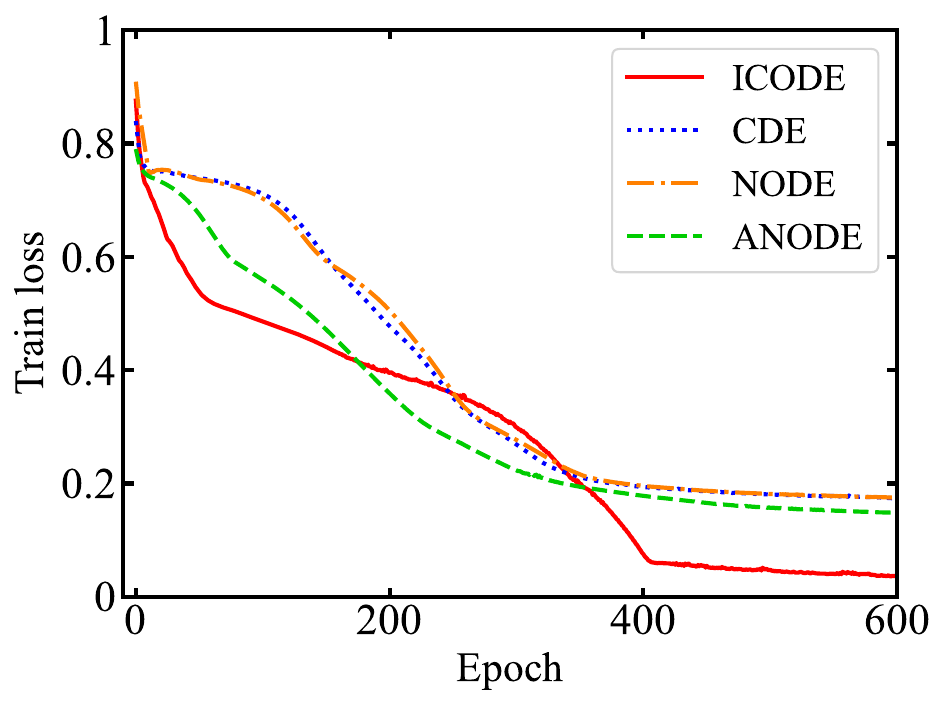}} 
 \subfloat[]{
		\label{Fig: DCDC Fig.sub.2}
		\includegraphics[width=0.33\textwidth]{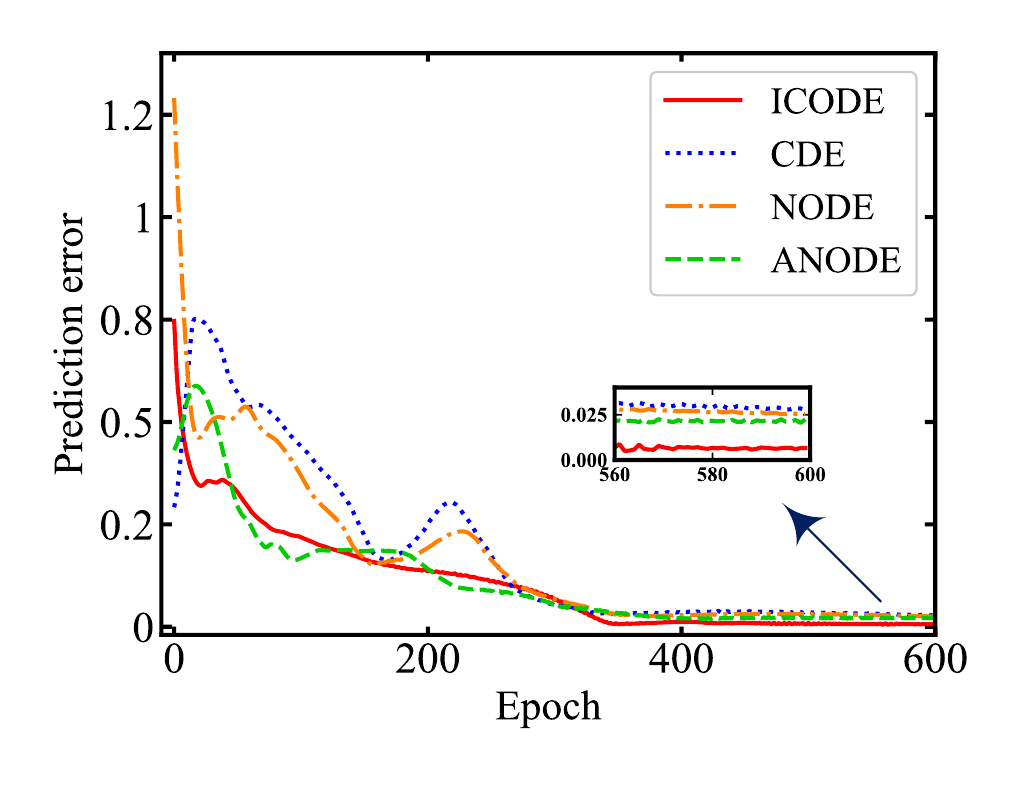}}
\subfloat[]{
		\label{Fig: DCDC Fig.sub.3}
		\includegraphics[width=0.32\textwidth]{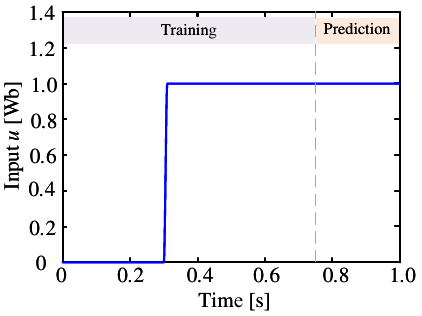}} \\
\subfloat[]{
    \label{Fig: DCDC Fig.sub.1-2}
    \includegraphics[width=0.33\textwidth]{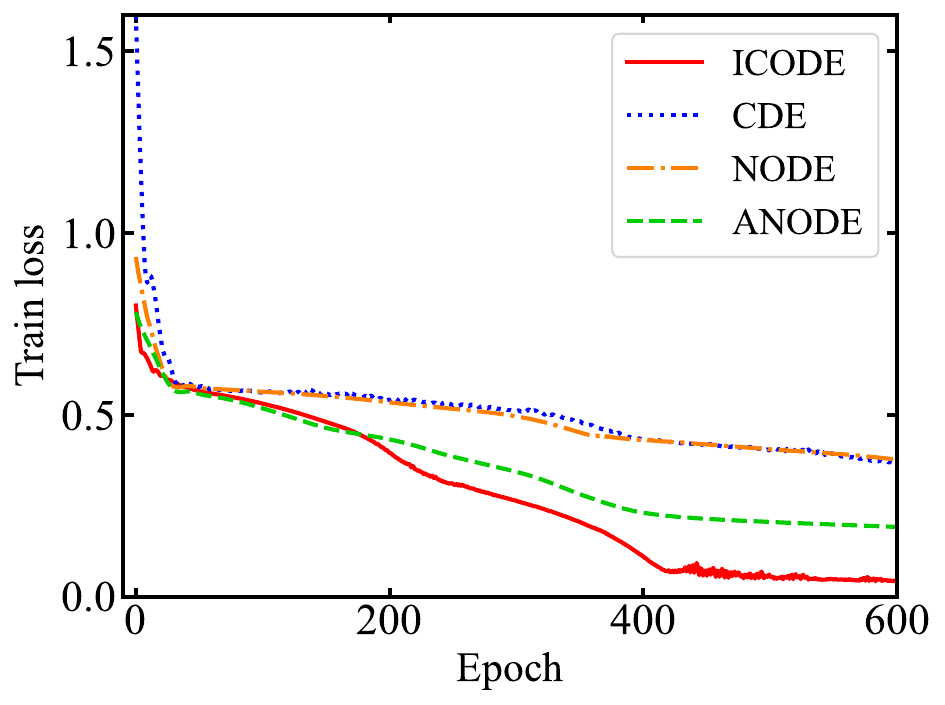}} 
 \subfloat[]{
		\label{Fig: DCDC Fig.sub.2-2}	\includegraphics[width=0.33\textwidth]{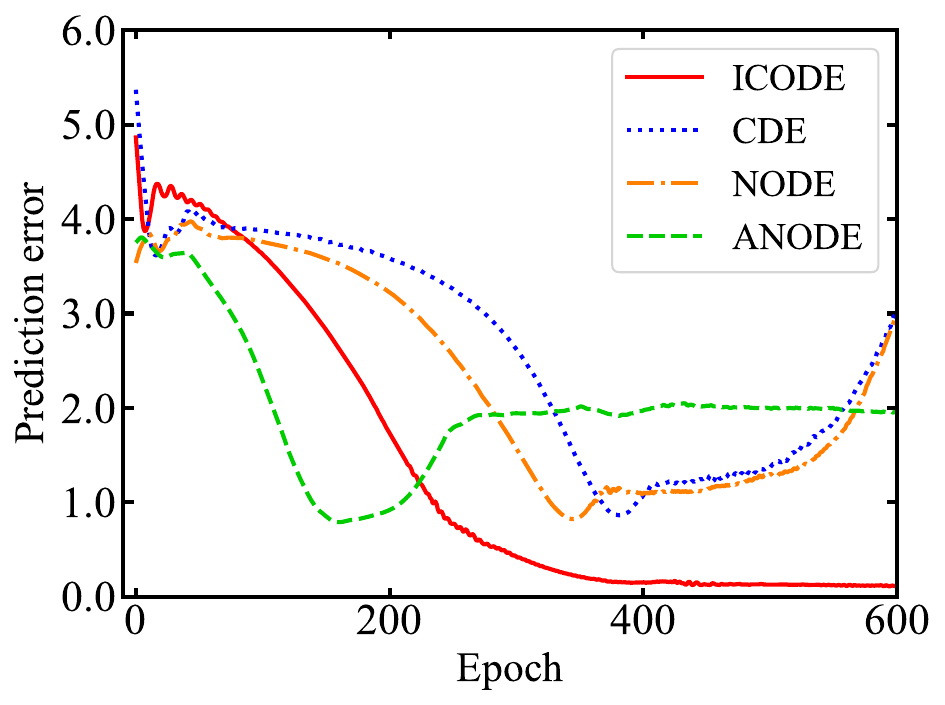}}
\subfloat[]{
		\label{Fig: DCDC Fig.sub.3-2}		\includegraphics[width=0.32\textwidth]{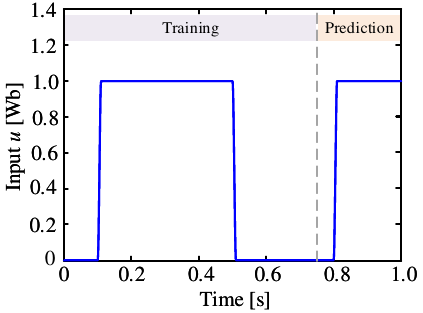}}
\caption{Comparison of training loss and prediction error across time steps for various models applied to the DC-to-DC converter task under two input conditions (top and bottom).}
\label{Fig: DCDC}
\end{figure*}

\emph{i)} In the first case, the control input changes its value from $0Wb$ to $1Wb$ during the training process, as presented in Fig. \ref{Fig: DCDC Fig.sub.3}. The corresponding training and prediction curves are depicted in Figs. \ref{Fig: DCDC Fig.sub.1}, \ref{Fig: DCDC Fig.sub.2}. Due to ICODE's ability to incorporate changes in $u$, it learns a more accurate model. Consequently, ICODE outperforms the other methods in prediction, achieving the smallest prediction error ($0.025$ in MSE), which is approximately 65\% of the error observed with CDE ($0.0383$ in MSE) and NODE ($0.0383$ in MSE). Though CDE can also detect variations in $u$ in this task, it receives an impulse response when $u$ switches between $0Wb$ and $1Wb$, and perceives no external input information if $u$ is unchanged. This behavior presents significant computational and optimization challenges for CDE in training. As the results indicate, in this setting, CDE's approach to incorporating external inputs is less effective than ICODE's. 

\emph{ii)} It is noteworthy that in the second scenario, the signal of control input $u$ jumps at $0.1s$, $0.5s$, and $0.8s$ (see Fig. \ref{Fig: DCDC Fig.sub.3-2} for details). This input is more complex, particularly with a switch occurring at $0.8s$, which falls within the testing phase. Neglecting this information could significantly impair prediction accuracy. As shown in Figs. \ref{Fig: DCDC Fig.sub.1-2}, \ref{Fig: DCDC Fig.sub.2-2}, ICODE consistently maintains strong performance in both learning and prediction. In contrast, other methods fail to further reduce prediction errors beyond 400 epochs, with continued training leading to severe overfitting.



\subsection{Motion of Rigid Body}
In this task, we deal with the angular momentum vector $x :=\begin{bmatrix} x_1 & x_2 & x_3 \end{bmatrix}^\top$ of a rigid body with arbitrary shape and mass distribution. The dynamics can be obtained using the fully actuated rigid body equations of motion \cite{chaturvedi2011rigid}, presented as follows. 
\begin{equation}\label{key7.23-4}
\begin{pmatrix}
	\dot{x}_1 \\ \dot{x}_2 \\ \dot{x}_3
\end{pmatrix}=
\begin{pmatrix}
	0 & -x_3 & x_2 \\
	x_3  & 0 & -x_1 \\
	-x_2 & x_1 & 0
\end{pmatrix}
\begin{pmatrix}
	x_1 / I_1 \\ x_2 / I_2 \\ x_3 / I_3
\end{pmatrix}+u.
\end{equation}
The coordinate axes are the principal axes of the body, with the origin of the coordinate system at the body's center of mass, and $I_1, I_2, I_3$ are the principal moments of inertia.

We trained on 10 trajectories with initial conditions 
$(\cos{(\phi)}, 0, \sin{(\phi)})$, where $\phi$ was sampled from a uniform distribution $\phi \sim \mathcal{U}(0.5, 1.5)$. For simplicity, the control input was set to be identical across the three channels, taking values from the set $[-1,1]$, with switching time instants at $0.1s$, $0.4s$, and $0.8s$. The total time interval is $[0, 1]$.

\begin{figure}
\centering
\includegraphics[width=0.4\textwidth]{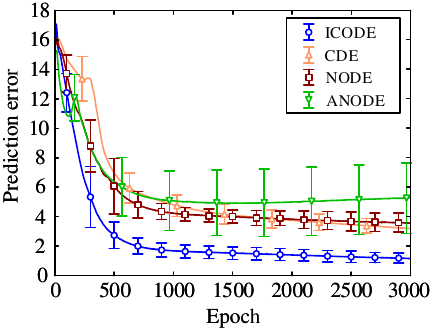}
\caption{Comparison of prediction errors at each time step on the test set for different models on the task of a rigid body.}
\label{Fig: rigBody}
 \end{figure}

In Fig~\ref{Fig: rigBody}, the lines stand for the prediction errors and the error bars mean the confidence intervals at each specific epoch. It reveals that ICODEs can achieve a very small prediction error with a high level of confidence.
This indicates that our proposed ICODEs can also effectively learn an operator $g$ in Eq.~\eqref{eq:CAS} such that $g(x)u = u$ for all $x$ and $u$, thereby ensuring the absence of coupling between the state and the input, as shown in the system~\eqref{key7.23-4}.

\begin{figure*}
	\centering
	\subfloat[]{
		\label{3D_Rigid_Fig.sub.1}
		\includegraphics[width=0.31\textwidth]{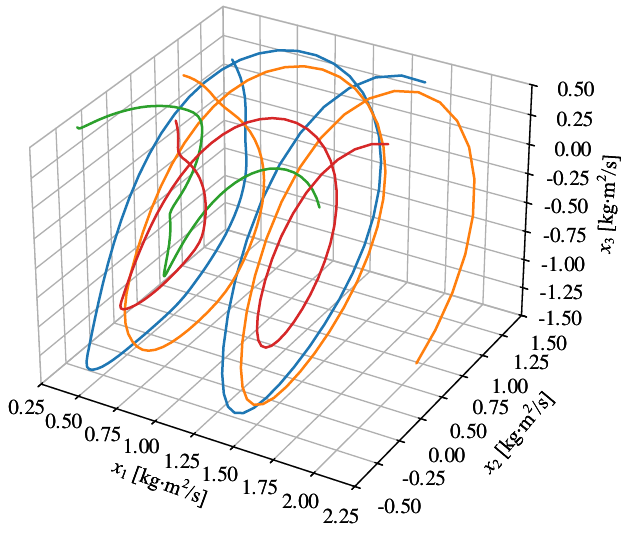}}  
  \subfloat[]{
		\label{3D_Rigid_Fig.sub.2}
		\includegraphics[width=0.31\textwidth]{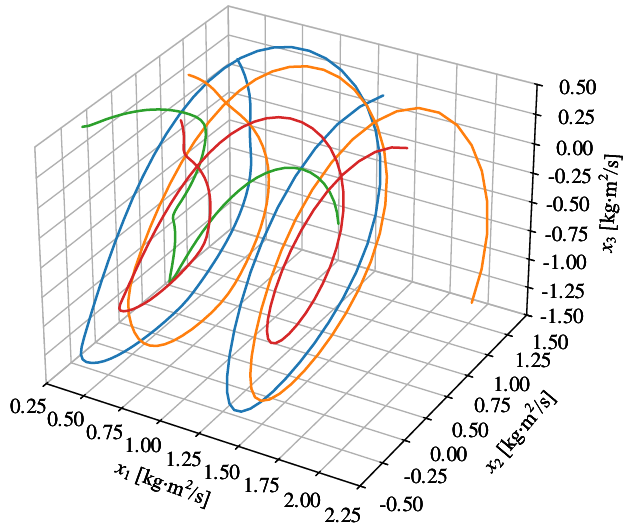}}
  \subfloat[]{
		\label{3D_Rigid_Fig.sub.3}	\includegraphics[width=0.31\textwidth]{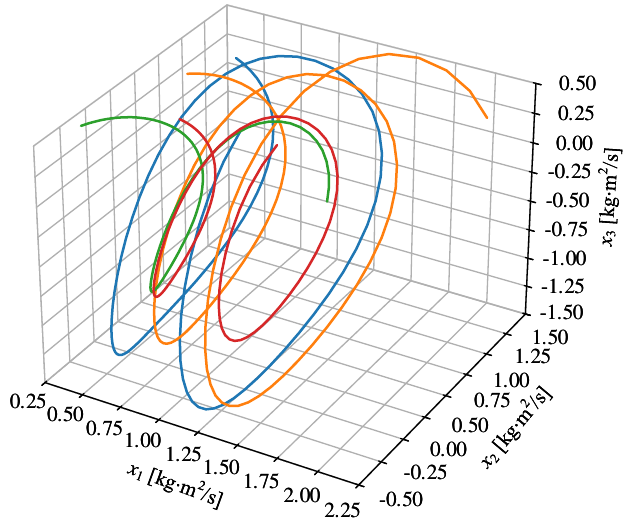}}
	\caption{The predicted trajectories of ICODE and NODE, compared with the ground truth on the task of a rigid body. (a) Ground Truth. (b) ICODE. (c) NODE. }
	\label{Fig: rigBody-Traj}
\end{figure*}

Fig. \ref{Fig: rigBody-Traj} illustrates the three-dimensional predicted trajectories of ICODE and NODE, in comparison with the ground truth. It is direct that ICODE provides a more detailed representation of the curves. Notably, ICODE accurately captures the abrupt twist at the tail of the red curve, a feature that the NODE fails to depict. This discrepancy appears mainly due to the lack of settings particularly dealing with external input in the NODE model.


\subsection{Rabinovich-Fabrikant Equation}
The Rabinovich-Fabrikant (R-F) equation  \cite{rabinovich1979stochastic} is a system of ordinary differential equations used to describe chaotic dynamics in some physical systems. Its form is given by
\begin{equation}\label{key7.23-6}
\begin{aligned}
\dot{x}&=y(z-1+x^2) + \gamma x, \\
\dot{y}&=x(3z+1-x^2)+\gamma y, \\
\dot{z}&=-2z(\alpha +xy).
\end{aligned}
\end{equation}
Here, the parameters $x,y,z$ are dimensionless, $\alpha$ and $\gamma$ are physical parameters that influence the dynamic behavior of the system. Suppose that there is a parameter drift in $\gamma$ and we can measure its value by corresponding sensors or soft sensing methods such as Kalman filtering or Gaussian processes.

\begin{figure*}
\centering
\subfloat[]{
		\label{Fig: RF1}
		\includegraphics[width=0.33\textwidth]{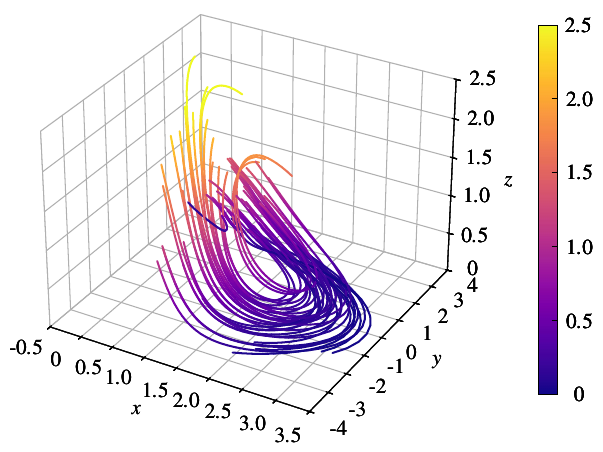}}
\subfloat[]{
		\label{Fig: RF2}		\includegraphics[width=0.33\textwidth]{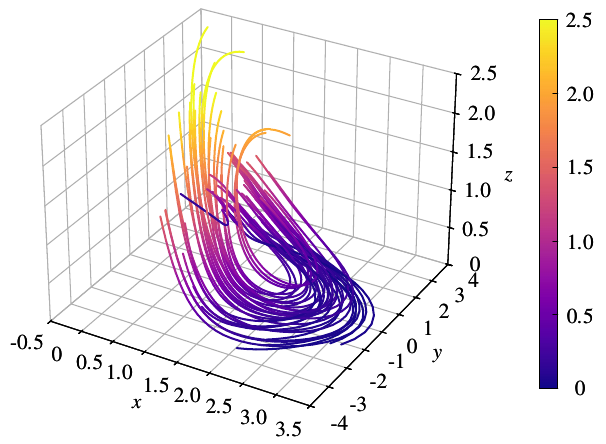}}
\subfloat[]{
		\label{Fig: RF3}		\includegraphics[width=0.33\textwidth]{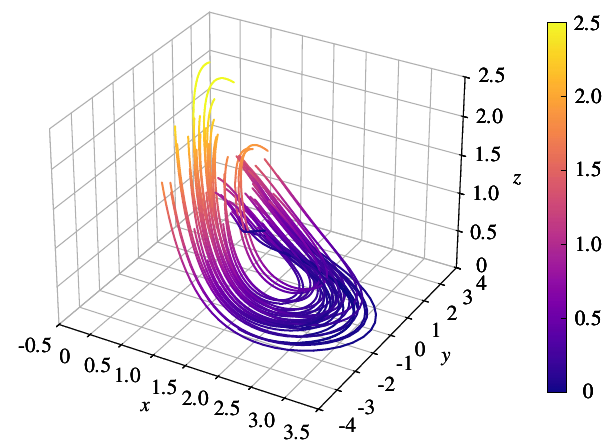}}
	\caption{Displays predictions by NODE and ICODE, for the R-F equation. (a) Ground truth. (b) ICODE. (c) NODE.}
	\label{Fig: combine_RF}
\end{figure*}

We trained on $40$ trajectories over a $1s$ interval with a time step: $\Delta t=0.02$. The values of $\gamma$ were gradually varied from $0.1$ to $1$ in time period $[0.2s, 1s]$. 
The visual results are presented in Fig. \ref{Fig: combine_RF}. Here we draw 40 trajectories and use different colors to distinguish points of different heights. We can see that ICODE predicts these trajectories more accurately, compared with the NODE.

\subsection{Glycolytic-glycogenolytic Pathway Model}

\begin{figure}[bh]
	\centering
		\includegraphics[width=0.5\textwidth]{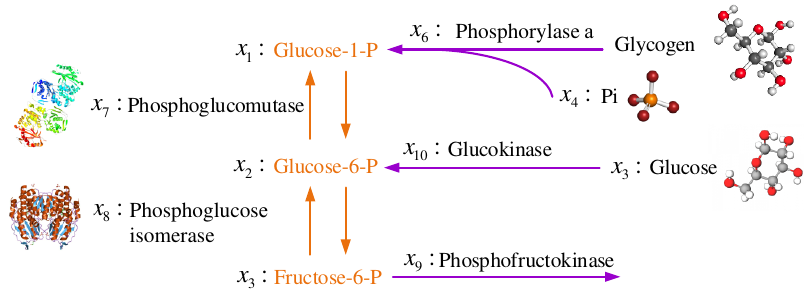}
	\caption{Diagram of the glycolytic-glycogenolytic pathway.}
	\label{Fig: Bio-machnisim}
\end{figure}

The glycolytic-glycogenolytic pathway has been extensively studied, with its biological mechanisms illustrated in Fig. \ref{Fig: Bio-machnisim} (refer to \cite{torres1994modelization, fnaiech2012intervention} for further details). Glycolysis refers to the metabolic process in which a single glucose molecule, comprising six carbon atoms, is enzymatically broken down into two molecules of pyruvate, each containing three carbon atoms. In contrast, glycogenolysis is the process by which glycogen, the stored form of glucose in the body, is degraded to release glucose when needed. This process is catalyzed by the enzyme phosphorylase, which sequentially cleaves glucose units from the glycogen polymer, producing glucose-1-phosphate as a product. The sequence of reactions in glycolysis and glycogenolysis can be modeled using an $S$-system framework, which is represented by
\begin{equation}\label{key:11.9-1}
\begin{aligned}
\dot{x}_1 &= \alpha_1 x_4^{\theta_{14}} x_6^{\theta_{16}} - \beta_1 x_1^{\mu_{11}} x_2^{\mu_{12}} x_7^{\mu_{17}}, \\
\dot{x}_2 &= \alpha_2 x_1^{\theta_{21}} x_2^{\theta_{22}} x_5^{\theta_{25}} x_7^{\theta_{27}} x_{10}^{\theta_{210}} - \beta_2 x_2^{\mu_{22}} x_3^{\mu_{23}} x_8^{\mu_{28}}, \\
\dot{x}_3 &= \alpha_3 x_2^{\theta_{32}} x_3^{\theta_{33}} x_8^{\theta_{38}} - \beta_3 x_3^{\mu_{33}} x_9^{\mu_{39}}.
\end{aligned}  
\end{equation}
The model variables are defined as follows: \( x_1 \) represents glucose-1-phosphate (glucose-1-P), \( x_2 \) represents glucose-6-phosphate (glucose-6-P), \( x_3 \) represents fructose-6-phosphate (fructose-6-P), \( x_4 \) represents inorganic phosphate (\( \text{Pi} \)), \( x_5 \) represents glucose, \( x_6 \) represents phosphorylase a, \( x_7 \) represents phosphoglucomutase, \( x_8 \) represents phosphoglucose isomerase, \( x_9 \) represents phosphofructokinase, and \( x_{10} \) represents glucokinase. In the context of the experiment, the parameters are specified as follows: \( \alpha_1 = 0.077884314 \), \( \theta_{14} = 0.66 \), \( \theta_{16} = 1 \), \( \beta_1 = 1.06270825 \), \( \mu_{11} = 1.53 \), \( \mu_{12} = -0.59 \), \( \mu_{17} = 1 \), \( \alpha_2 = 0.585012402 \), \( \theta_{21} = 0.95 \), \( \theta_{22} = -0.41 \), \( \theta_{25} = 0.32 \), \( \theta_{27} = 0.62 \), \( \theta_{210} = 0.38 \), \( \beta_2 = \alpha_3 = 0.0007934561 \), \( \mu_{22} = \theta_{32} = 3.97 \), \( \mu_{23} = \theta_{33} = -3.06 \), \( \mu_{28} = \theta_{38} = 1 \), \( \beta_3 = 1.05880847 \), \( \mu_{33} = 0.3 \), and \( \mu_{39} = 1 \). The independent variables \( x_4 \), \( x_5 \), and \( x_8 \) are treated as manipulated variables, which influence the system's dynamic behavior. Specifically, the system is described by the equations \( \dot{x}_4 = u_1 \), \( \dot{x}_5 = u_2 \), and \( \dot{x}_6 = u_3 \), where \( u_1 \), \( u_2 \), and \( u_3 \) represent external inputs.

To better capture the complexities of these dynamics, we introduce adaptations to existing methods. Recently, NODE and its variants have been combined with Transformer layers (see, for example, \cite{zhongneural,mei2024controlsynth}). Integrating Transformer architectures has demonstrated substantial potential, proving to be a highly effective tool; thus, we adopt four NODE-based methods by incorporating Transformer layers (naming convention: “Model-Adapt”; see the second group of Table \ref{tab: Bio}). The prediction outcomes are presented in Table \ref{tab: Bio}, with values in parentheses ($\pm \cdot$) indicating the standard deviation. Among the evaluated methods, ICODE consistently achieves the highest performance, both with and without Transformer layers. Furthermore, the inclusion of Transformer layers significantly enhances ICODE's predictive accuracy, thanks to the Transformer’s robust capacity to leverage external information. This improvement, however, is less pronounced in the other methods.

\begin{table}[H]
\small
\caption{ Prediction results of four models and their variants (integrated with Transformer layers) on the dynamics of the glycolytic-glycogenolytic pathway. Best performing models are marked in \textcolor[rgb]{0.25, 0.5, 0.75}{\bfseries{blue}}, second-best in \textcolor[rgb]{0.75, 0.35, 0.1}{\bfseries{brown}}.}
\centering
\scalebox{0.9}{
    \begin{tabular}{l|cc}
        \toprule
        \textbf{Model} & \textbf{MSE} & \textbf{MAE}  \\
        \midrule
        ICODE (Ours) & \textcolor[rgb]{0.75, 0.35, 0.1}{\bfseries{0.0496}} ($\pm 0.0191$) & \textcolor[rgb]{0.75, 0.35, 0.1}{\bfseries{0.623}} ($\pm 0.0797$)  \\
        CDE & 0.0912 ($\pm 0.00628$) & 1.026 ($\pm 0.0308$)  \\
        NODE & 0.0632 ($\pm 0.00512$) & 0.847 ($\pm 0.0287$)  \\
        ANODE & 0.0593 ($\pm 0.00603$) & 0.831 ($\pm 0.0389$)  \\
        \midrule
        ICODE-Adapt  (Ours) &  \textcolor[rgb]{0.25, 0.5, 0.75}{\bfseries{0.00721}} ($\pm  0.00238$) & \textcolor[rgb]{0.25, 0.5, 0.75}{\bfseries{0.295}} ($\pm 0.0433$)  \\
        CDE-Adapt & 0.0854 ($\pm  0.0144$) & 0.856 ($\pm  0.0829$)  \\
        NODE-Adapt & 0.128 ($\pm  0.0214$) & 0.995 ($\pm 0.0877$) \\
        ANODE-Adapt & 0.0609 ($\pm 0.00659$) & 0.837 ($\pm 0.0392$)  \\
        \bottomrule
    \end{tabular}
}
\label{tab: Bio}
\end{table}

\subsection{Swing Equation}

In this example, we analyze the performance of ICODE in multi-agent cooperative systems by considering the swing equation \cite{kundur2007power}, which models the dynamics of synchronous generators or motors in an electrical power system. The dynamics of each node are typically described by the following second-order differential equation:
\begin{equation}
M_i \frac{d^2\theta_i}{dt^2} + D_i \frac{d\theta_i}{dt} = P_i - \sum_{j=1}^{N} K_{ij} \sin(\theta_i - \theta_j),
\end{equation}
where $M_i$ represents the moment of inertia of generator $i$, $D_i$ is the damping coefficient accounting for the system's frequency damping effects, and $\theta_i$ is the angular position (phase angle) of generator $i$. The term $P_i$ denotes the power input to generator $i$, and $K_{ij}$ is the electrical power transmission coefficient between nodes $i$ and $j$, which is related to the electrical conductance and link parameters of the transmission line. The term $\sin(\theta_i - \theta_j)$ represents the power flow between nodes.
To convert this system to a first-order differential equation form, we introduce state variables $\omega_i:= \frac{d\theta_i}{dt}$, representing the angular velocity. Consequently, the state dynamics of each node can be expressed as the following two sets of first-order differential equations:
$\frac{d\theta_i}{dt} = \omega_i, \; \frac{d\omega_i}{dt} = \frac{1}{M_i} \left( P_i - \sum_{j=1}^{N} K_{ij} \sin(\theta_i - \theta_j) - D_i \omega_i \right).$
We consider a network of 10 nodes. The corresponding topology graph is provided in appendix B. The simulation is conducted over the time interval \([0s, 5s]\). The external input \( P_i \) follows the same pattern as depicted in Fig. \ref{Fig: DCDC Fig.sub.3-2}, varying between 0 and 1 at \( t = 0.5s, 2.5s, \) and \( 4.5s \). The parameters \( M_i \) and \( D_i \) are independently sampled from uniform distributions over the intervals \([0.3, 0.9]\) and \([0.7, 1.3]\), respectively. A total of 128 trajectories are generated and partitioned into batches, each containing 16 trajectories. In the ANODE framework, we augment the system with 50 additional state variables.

\begin{figure}[htbp]
    \centering
    \subfloat{%
        \includegraphics[width=0.4\textwidth]{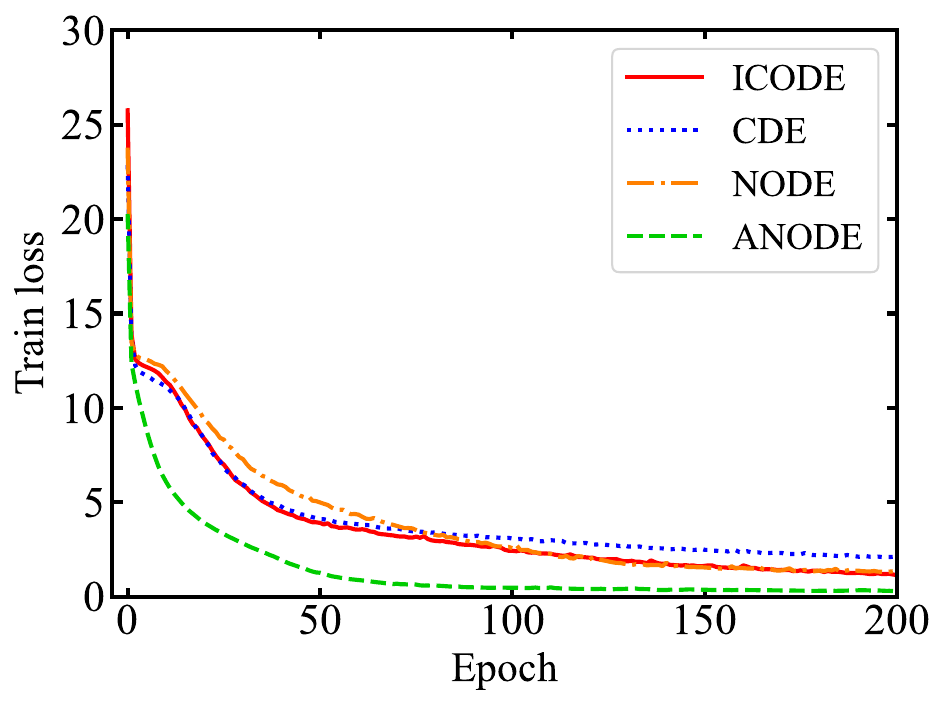} 
        \label{network-trainError}
    } \\[-0.5cm]
    
    \subfloat{%
        \includegraphics[width=0.4\textwidth]{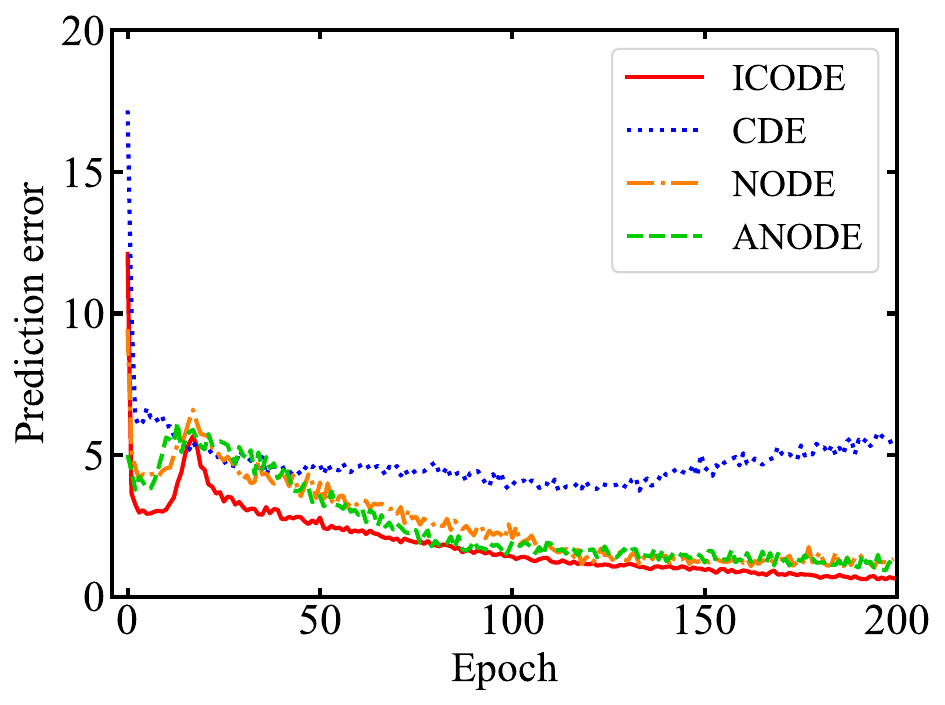} 
        \label{fig:network-testError}
    }

    \caption{Comparison of training loss and prediction error across time steps for various models applied to the Swing Equation.}
    \label{fig:network}
\end{figure}

The results are presented in Fig. \ref{fig:network}. It is clear that ANODE achieves the lowest training loss, which is expected given that the augmented states enhance the network's degrees of freedom and fitting capacity. However, the prediction error curves indicate that this improved fitting ability does not translate well to generalization. A similar issue is observed with CDE, as its structure does not adapt effectively to these models. In contrast, ICODE consistently demonstrates superior performance in this scenario.


\subsection{Heat Conduction Equation}

The heat conduction equation describes how heat diffuses through a given region over time. It is a fundamental equation in the study of heat transfer and has been widely used in physics, engineering, and applied mathematics. The general form of a nonlinear heat conduction equation can be written as
\begin{equation*}
	\frac{\partial{T}}{\partial{t}} =  \nabla \cdot (k(T) \nabla T ) + Q(T, \nabla T),
\end{equation*}
where  $T(t, x) \in \mathbb{R}$ represents the temperature as a function of time $t$ and spatial coordinate vector $x$ (here, $x_i \in [0,10]$ in this setting), $k(T)$ denotes the thermal conductivity, which is temperature-dependent, $\nabla$ is the gradient operator \emph{w.r.t.} the spatial vector,
and 
$Q(T, \nabla T)$ accounts for internal heat sources or sinks that may depend on temperature and its gradient. We assume that the heat source  $Q(\cdot, \cdot)$ is known and treat it as the input to the model. 
\subsubsection{One-Dimensional Case} In this scenario, the spatial domain is one-dimensional, meaning $T(t, \cdot) \in C(\mathbb{R} \times \mathbb{R}, \mathbb{R})$, where the value of $t$ is fixed. We assume the absence of internal heat sources, with only boundary values specified, corresponds to an initial boundary value problem for the heat conduction equation. Initial values are randomly generated within the interval $[-1, 1]$, and the boundary values are defined by $T(t, x) = 2\sin(2\pi t)\exp(-t/5) + 0.1$ for $x \in \{0, 10\}$. This known boundary information is regarded as extrinsic input.

\begin{figure}
	\centering	
		\includegraphics[width=0.45\textwidth]{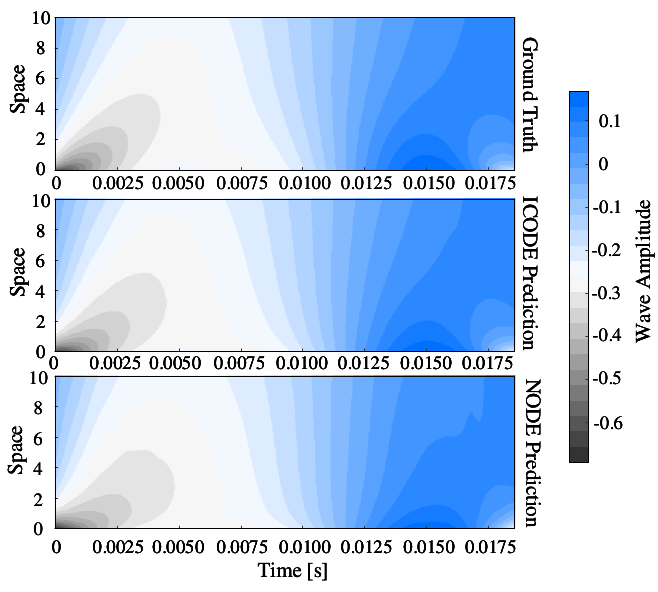}
	\caption{Visualized prediction results of one-dimensional heat conduction equation.}
	\label{Fig: heat-1dim-time-space}
\end{figure}

The prediction performance evaluated by MSE is presented in Table \ref{tab: pde-1dim}, with values in parentheses indicating the standard deviation (calculated over 10 trials). ICODEs achieve the highest performance, followed by CDEs. ANODEs exhibit the lowest performance, mainly due to the increased network complexity introduced by augmented states and networks, which complicates training and diverges from the original requirements, as the heat conduction equation does not necessitate additional latent variables for effective representation. Fig. \ref{Fig: heat-1dim-time-space} provides visual comparisons of ICODEs and NODEs at various spatio-temporal points against the ground truth, illustrating ICODEs' superior long-term predictive accuracy.

\begin{table}[H]
\small
\caption{ MSE prediction results of one-dimensional heat conduction equation.}
\scalebox{0.9}{
\begin{tabular}{clcl}
\toprule
ICODE: &  \textbf{0.00245} ($\pm 0.000923$)   & 
CDE: &  0.00276($\pm 0.000435$)   \\
NODE:&  0.00319 ($\pm 0.00118$)  &
ANODE: & 0.0437 ($\pm 0.01064$) \\
\bottomrule
\end{tabular}}
\label{tab: pde-1dim}
\end{table}

\begin{table}[H]
\small
\caption{ Final prediction error and the standard variance in the two-dimensional heat conduction equation.}
\centering
\scalebox{0.9}{
\begin{tabular}{clcl}
\toprule
ICODE: &  \textbf{0.00172} ($\pm 0.00011$)   & 
CDE: &   0.00178  ($\pm 9.74e-5 $)   \\
NODE:&  0.00177 ($\pm 9.87e-5$)   &
ANODE: & 0.0201 ($\pm 0.0011 $)  \\
\bottomrule
\end{tabular}}
\label{tab: pde-2dim}
\end{table}

\subsubsection{Two-Dimensional Case}  In this scenario, the spatial domain is two-dimensional, that is, $T(t, \cdot) \in C(\mathbb{R} \times \mathbb{R}^2, \mathbb{R})$ (the second independent variable $x$ is two-dimensional). It is assumed that \( Q = 10 \) during the intervals spanning 10\%–40\% and 60\%–90\% of the total time period, with \( Q \) set to zero outside these ranges. The final predicted results are presented in Table \ref{tab: pde-2dim}. The results indicate that ICODEs exhibit the smallest prediction error. Conversely, ANODEs demonstrate the poorest performance, primarily due to significant overfitting.


\begin{figure*}
	\centering
	\subfloat[]{
		\includegraphics[width=0.31\textwidth]{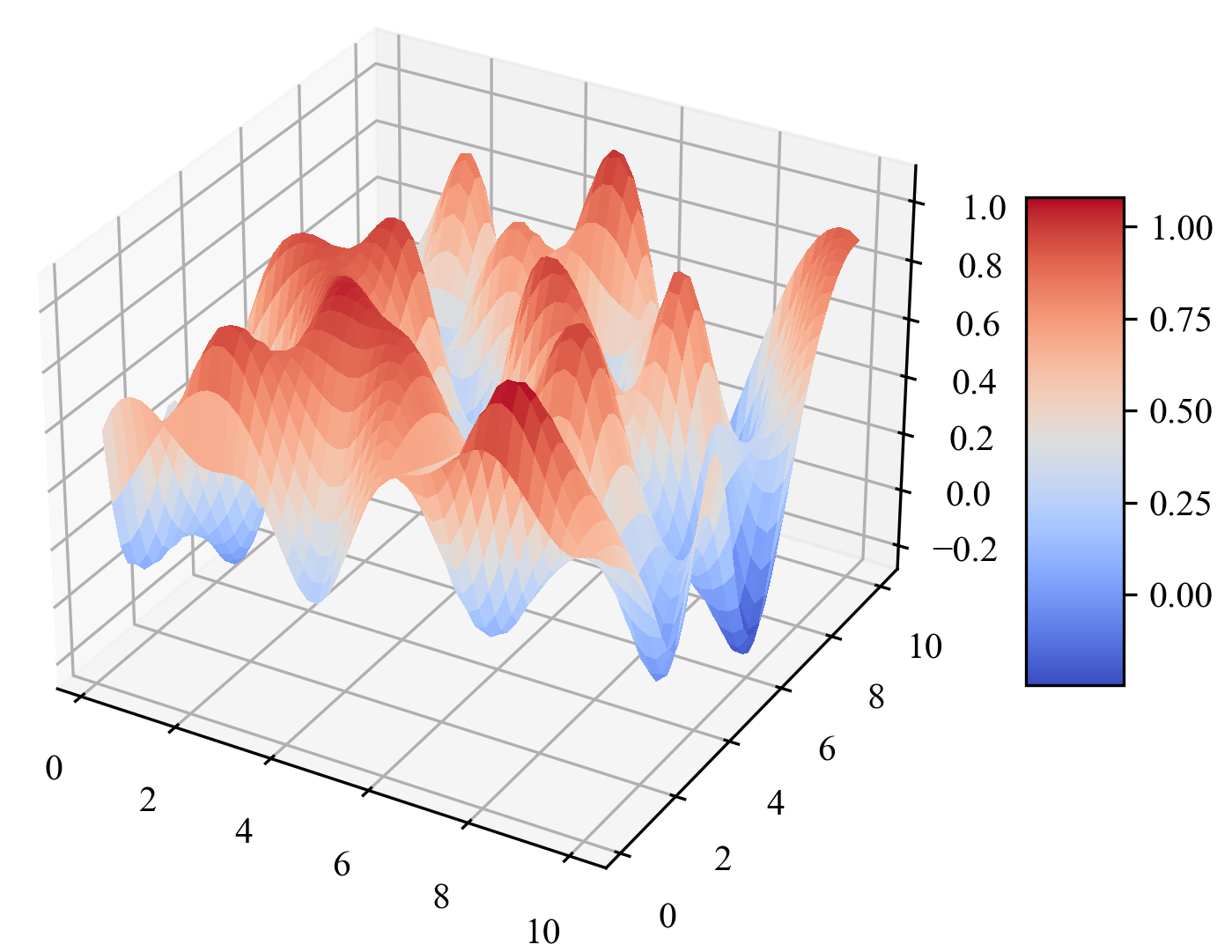}}   \label{Fig: heat-3D_a}
	\subfloat[]{
		\includegraphics[width=0.3\textwidth]{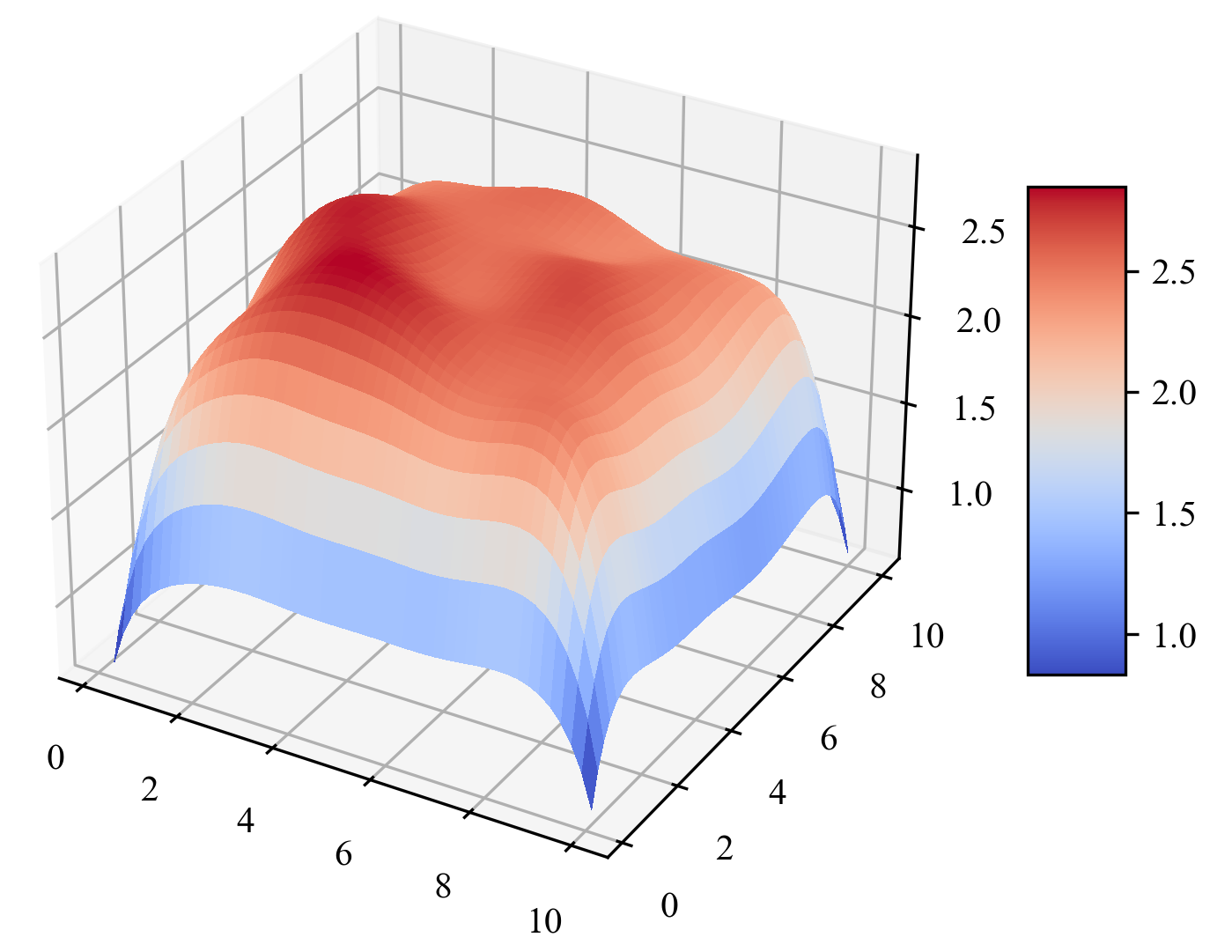}}  \label{Fig: heat-3D_b}
	\subfloat[]{
		\includegraphics[width=0.3\textwidth]{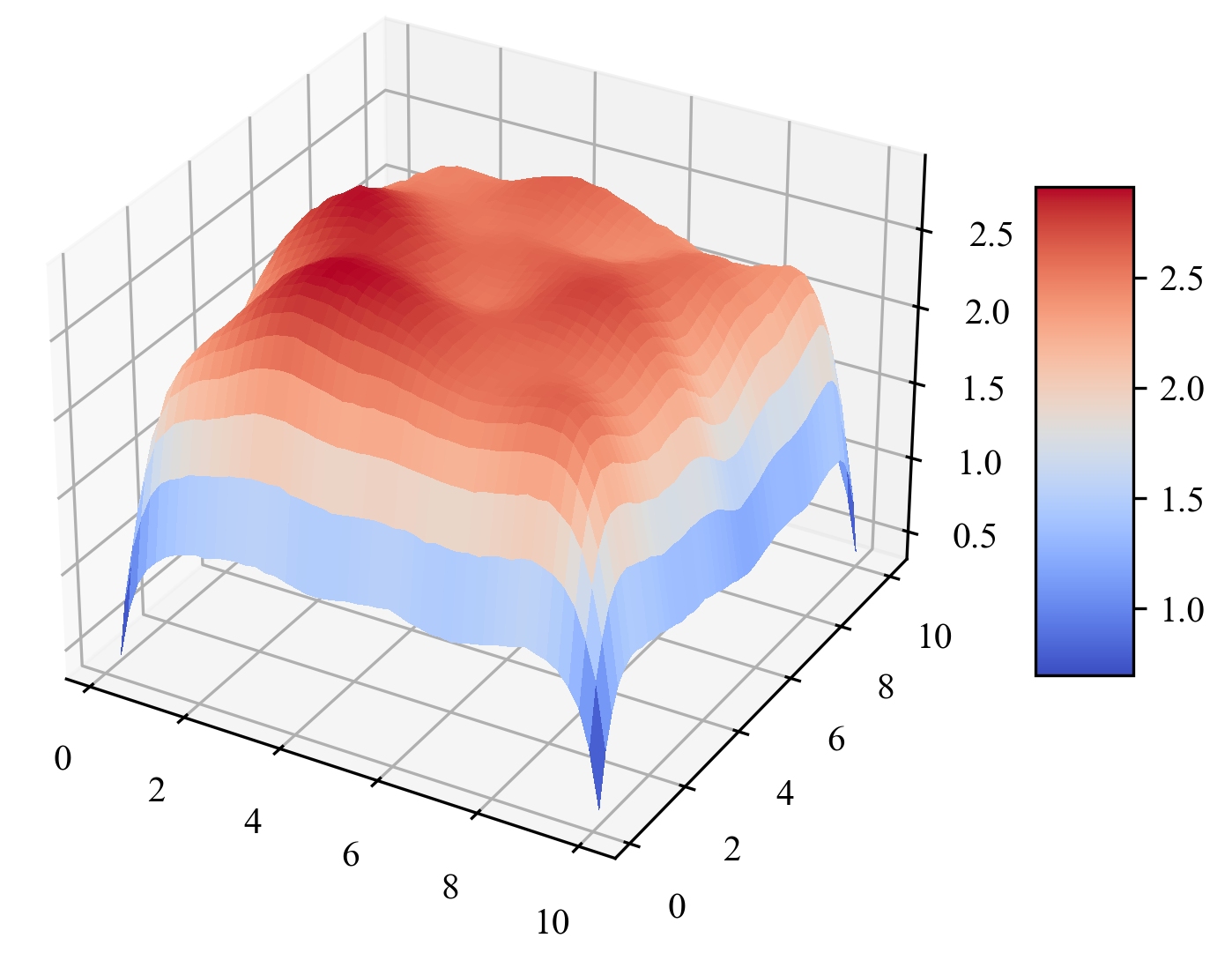}}   \label{Fig: heat-3D_c}
	\caption{Displays initial conditions and prediction by ICODE for heat conduction equation, compared with the ground truth. (a) Initial Condition. (b) Ground Truth. (c) Prediction by ICODE.}
	\label{Fig: heat-3D}
\end{figure*}

Figure \ref{Fig: heat-3D} illustrates the initial condition, ground truth, and predictions generated by ICODE. The figure reveals a substantial increase in surface temperature due to the heat source \( Q \). ICODE effectively captures both the large-scale temperature variations and the finer details, including the raised and depressed regions depicted in the visualization. These findings highlight the robust modeling capabilities of ICODE.

\section{Further Statistical Studies} \label{sec:statistical}
This section details the statistical methods utilized to ensure the reliability and scientific validity of our experimental results. Specifically, it emphasizes the calculation of standard deviation percentages and the performance in noisy environments as indicators of variability and consistency across multiple experimental runs.

\begin{figure}
\centering	
\includegraphics[width=0.43\textwidth]{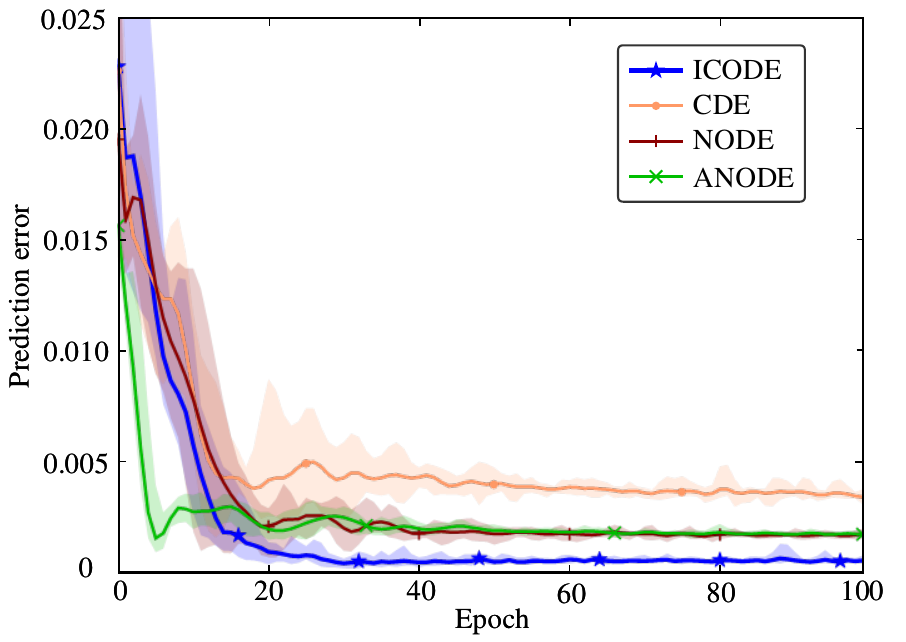}
\caption{Comparison of prediction errors with 95\% confidence intervals at each time step on the test set for different models on the task of the single-link robot.}
\label{Fig: slrobot-std}
\end{figure}

We conducted additional experiments on the single-link robot in the third scenario ($\Delta u=0.2$), repeating the experiment $10$ times. The mean MSE loss curve for the prediction phase is illustrated in Fig. \ref{Fig: slrobot-std}, with the shaded area representing the 95\% confidence intervals of every single experiment for each method. The corresponding MSE and MAE values at the final point (epoch 100) are detailed in Table \ref{tab: slrobot-std}. Here numbers inside brackets indicate the standard variation.
The results indicate that ICODE consistently demonstrates superior and steady performance.

\begin{table}[H]
\small
\caption{ Final prediction error and the standard variance in the single-link robot experiment.}
\centering
\scalebox{0.9}{
    \begin{tabular}{l|cc}
        \toprule
        Model & MSE & MAE   \\
        \midrule
        ICODE (Ours) & \textbf{0.025}($\pm 0.012$) & \textbf{0.060} ($\pm 0.0071$)  \\
        CDE & 0.059  ($\pm 0.011 $) & 0.130 ($\pm 0.0051$)  \\
        NODE & 0.042 ($\pm 0.0089$) & 0.100 ($\pm 0.0026$)  \\
        \bottomrule
    \end{tabular}
}
\label{tab: slrobot-std}
\end{table}

\begin{figure}
	\centering	
		\includegraphics[width=0.43\textwidth]{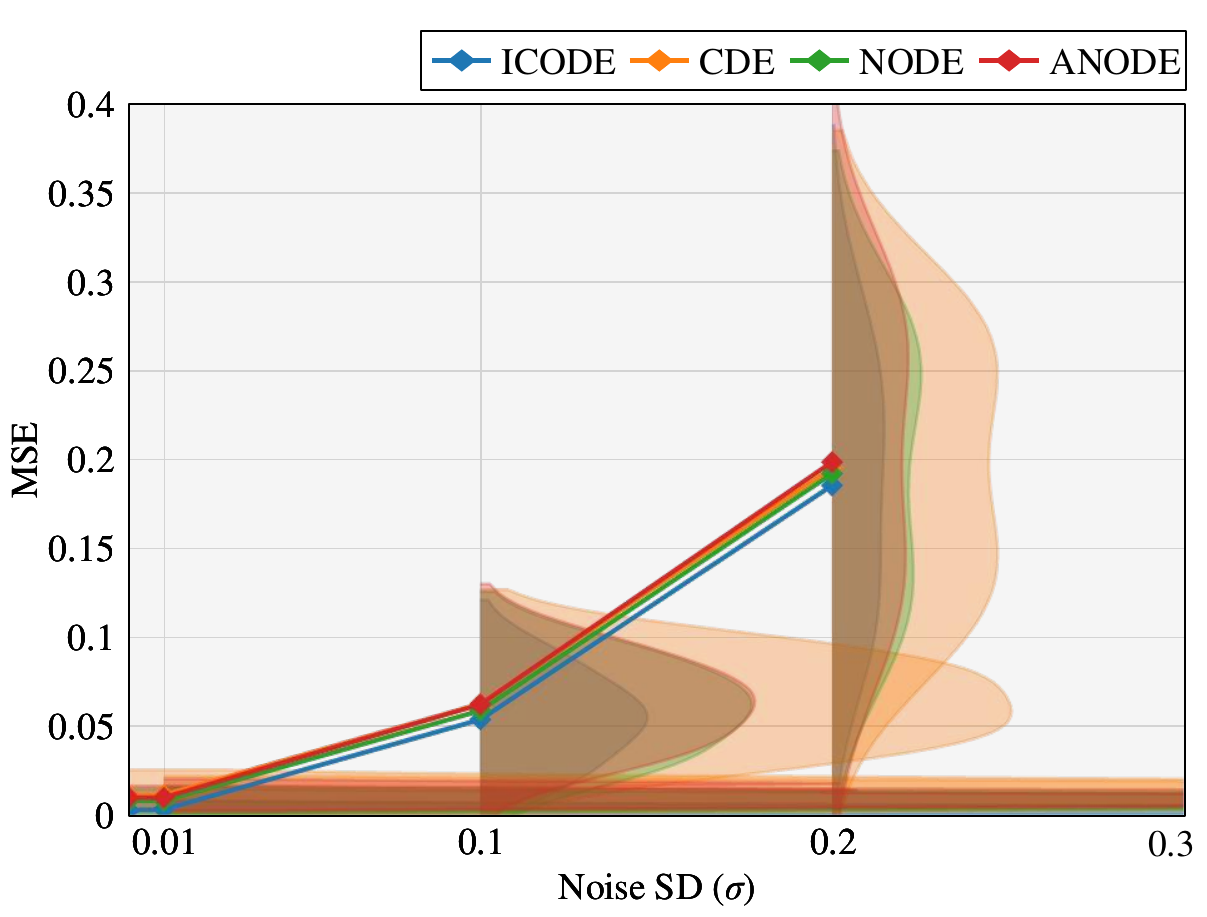}
	\caption{MSE comparison of methods on the task of the single-link robot under Gaussian noise with different magnitudes. Lines show mean MSE; shaded areas indicate MSE distribution.}
	\label{Fig: noise}
\end{figure}

To further explore the robustness of these methods in noisy environments, we added Gaussian noise of different magnitudes to the training data. The prediction results are drawn in Fig. \ref{Fig: noise}, with numerical details in Table \ref{tab: noise}, where $p$ indicates the noise level. 
In Fig. \ref{Fig: noise}, the lines represent the average MSE prediction loss, and the shaded areas capture the distribution of loss in these three cases ($p=0.01,0.1,0.2$).
This figure illustrates that noises will impact the performance of these methods, increasing the prediction errors and uncertainty (greater variance). However, whichever the case, ICODEs remain the best performer. This demonstrates the superiority of ICODEs in terms of robustness. 

\begin{figure}
\centering	
\includegraphics[width=0.43\textwidth]{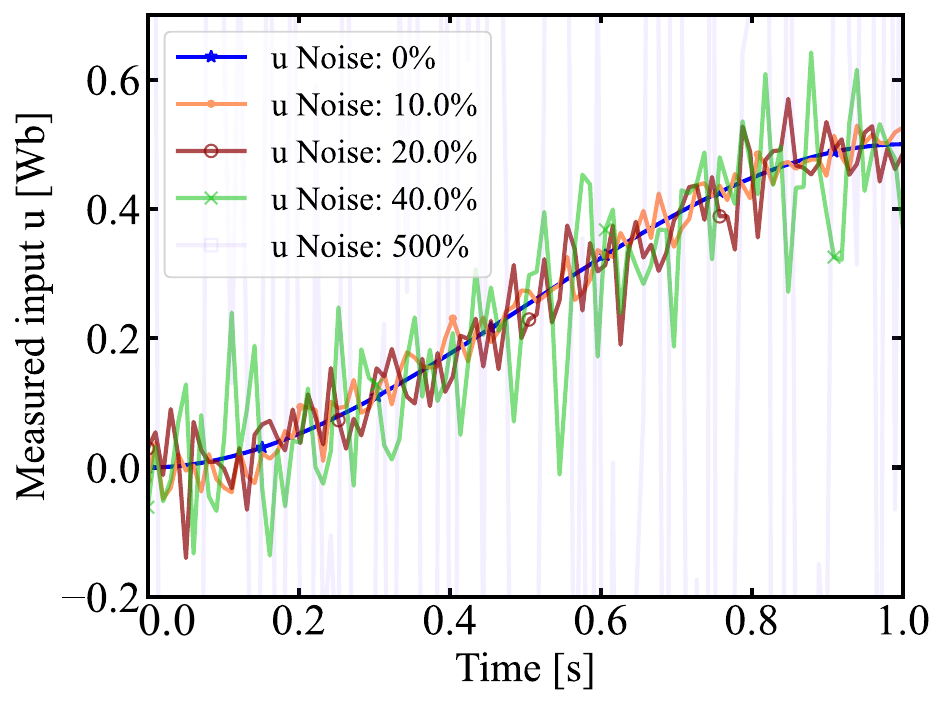}
\caption{The input $u$ with measurement noises of different magnitudes.}
\label{Fig: compare_U_noiseU}
\end{figure}

\begin{figure}
\centering	
\includegraphics[width=0.43\textwidth]{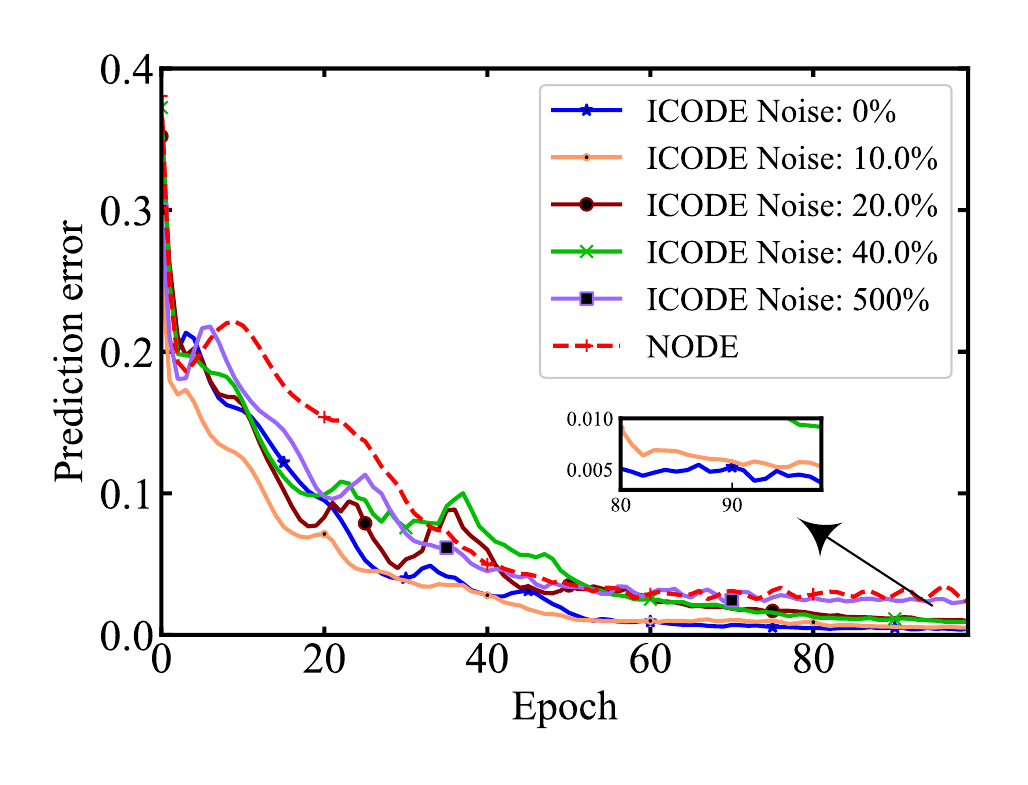}
\caption{Prediction errors on the test set for ICODE under different measurement noises in $u$, compared with NODE.}
\label{Fig: compare_noiseU}
\end{figure}

We also investigated the scenario in which the measured input \( u \) is affected by measurement noise. To simulate this, we introduced Gaussian noise of varying magnitudes into the measured \( u \), while ensuring that the system trajectories are generated without these disturbances. Specifically, noise was added at proportions of \( 10\% \), \( 20\% \), \( 40\% \), and \( 500\% \). The resulting noisy input signals are illustrated in Fig. \ref{Fig: compare_U_noiseU}, with the corresponding prediction errors shown in Fig. \ref{Fig: compare_noiseU}. 
 
As expected, ICODE achieves the best performance when \( u \) is noise-free (represented by the blue curve). When a small amount of noise is introduced, the prediction error remains largely unaffected. Under extreme noise conditions (\( 500\% \)), where the external input \( u \) carries little useful information, ICODE's performance degrades to a level comparable to that of NODE. Notably, despite the presence of significant disturbances in the input signal, ICODE remains effective. These results highlight ICODE’s robustness to noise in the input \( u \).

\begin{table*}
\small
\caption{Performance comparison under various amounts of noises. } 
\centering
\renewcommand{\arraystretch}{1.2}
\setlength{\tabcolsep}{2.5mm}{
\scalebox{0.9}{
\begin{tabular}{cccccccccccc}
\toprule
\cmidrule[0.8pt]{1-9} 
 \multicolumn{3}{c|}{\multirow{2}{*}{Model}}       
 & \multicolumn{2}{c|}{$p=0.01$}              & \multicolumn{2}{c|}{$p =0.1$}   & \multicolumn{2}{c}{$p=0.2$}                                   \\ 
\multicolumn{3}{c|}{} 
& MSE     & \multicolumn{1}{c|}{MAE}   & MSE     & \multicolumn{1}{c|}{MAE}    & MSE     & \multicolumn{1}{c}{MAE}   \\ 
\midrule
\multicolumn{3}{c|}{ICODE (Ours)}          & \textbf{0.00326} ($\pm 0.00137$)       & \multicolumn{1}{c|}{\textbf{0.182} ($\pm 0.0382$)}                   & \multicolumn{1}{c}{\textbf{0.0538} ($\pm 0.0229$)}              & \multicolumn{1}{c|}{\textbf{0.756} ($\pm 0.155$)} & \textbf{0.185 ($\pm 0.067$)} &  \multicolumn{1}{c}{\textbf{1.389} ($\pm 0.276$)}    \\

\multicolumn{3}{c|}{CDE}             & 0.0113 ($\pm 0.00386$)     & \multicolumn{1}{c|}{0.351 ($\pm 0.0685$)}                   & \multicolumn{1}{c}{ 0.0626 ($\pm 0.0223$)}                & \multicolumn{1}{c|}{0.831 ($\pm 0.150$)}  &  0.195 ($\pm 0.0690$) & \multicolumn{1}{c}{1.435 ($\pm 0.260$)}    \\

\multicolumn{3}{c|}{NODE}             &  0.00810 ($\pm 0.00259$)     & \multicolumn{1}{c|}{0.299 ($\pm 0.0487$)}                   & \multicolumn{1}{c}{0.0589 ($\pm 0.0232$)}                & \multicolumn{1}{c|}{ 0.801 ($\pm 0.156$)}  &  0.192 ($\pm 0.0689$) & \multicolumn{1}{c}{1.418 ($\pm 0.261$)} \\

\multicolumn{3}{c|}{ANODE}             & 0.00996 ($\pm 0.00354$)    & \multicolumn{1}{c|}{0.332 ($\pm 0.0682$)}                    & \multicolumn{1}{c}{0.0623 ($\pm 0.0221$) }             & \multicolumn{1}{c|}{0.829 ($\pm 0.142$)}  &  0.198 ($\pm 0.0753$) & \multicolumn{1}{c}{1.446 ($\pm 0.268$)} \\                       
\bottomrule
\end{tabular}}}
\label{tab: noise}
\end{table*}

\begin{figure*}
\centering
\subfloat[]{
		\label{Fig: differentU}
		\includegraphics[width=0.32\textwidth]{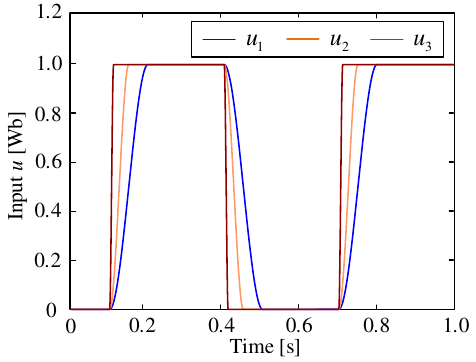}}
\subfloat[]{
		\label{Fig: scatter}	
        \includegraphics[width=0.35\textwidth, trim=0 5 0 0, clip]{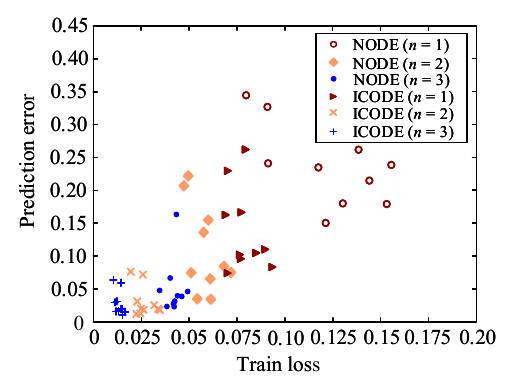}}
\subfloat[]{
		\label{Fig: heatmap}		\includegraphics[width=0.32\textwidth]{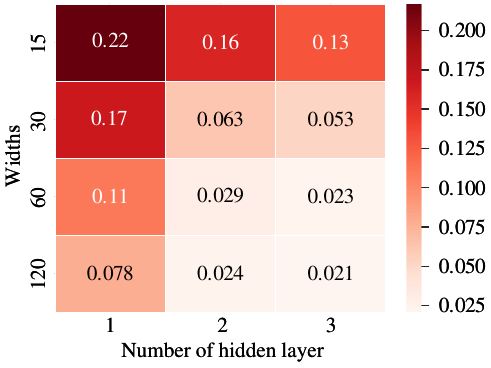}}
	\caption{The shapes of different piecewise inputs (left). Relation between training loss and prediction error for various numbers of hidden layers at width 60 (middle).  Final validation losses for different configurations of ICODE (right).}
 
	\label{Fig: combine}
\end{figure*}

\section{Experiments on Input Role}
In this section, we conduct in-depth examinations of the effect of inputs on the learning of dynamical systems. First, we design an experiment to investigate the importance of the external input. More specifically, we gradually increase the value of the input and analyze the response of our model. Moreover, we study the influence of the shapes of inputs on the models. We demonstrate the advantage of ICODE over the existing CDE method, highlighting its enhanced capability to incorporate incoming information. This results in higher performance, particularly when handling atypical inputs, compared to CDE and other selected NODEs.

\subsection{Input Span}
To examine the significant role of the input signal $u$, we conduct an experimental study comparing the performance of different methods under increasing values of $u$. Using the single-link robot and rigid body models as tasks, we vary the value of the control term $u$ within the range $[-k_u, k_u]$. When  $k_u$ equals 0, there is no input to the system, while larger values of $k_u$ indicate greater external influences on the dynamical system. Table \ref{tab:comparek_u} presents the training results for various $k_u$, where RMSE stands for Root Mean Square Error. When $k_u=0$, the performance of our method reduces to that of the standard NODE, showing no superiority over other methods. With $k_u=0.2$, the model experiences a small external influence, leading to larger errors in RMSE and MAE for NODE and ANODE, as they lack mechanisms to handle this variation. In this scenario, our method begins to demonstrate advantages in learning and prediction. When $k_u=1$, indicating a conspicuous external influence, the prediction error of our method is markedly lower than that of other methods.

\begin{table*}[h!]
\small
\caption{Performance comparison of different methods in the three scenarios of input span.}
\centering
\renewcommand{\arraystretch}{1.2}
\setlength{\tabcolsep}{2.1mm}{
\begin{tabular}{ccccc|cccc|cccc|ccc}
\toprule
\cmidrule[0.8pt]{1-15} & \multicolumn{2}{c|}{}  
 & \multicolumn{6}{c|}{Single-link Robot}                                 & \multicolumn{6}{c}{Rigid Body}      \\ 

 \cmidrule{4-15} 
\multicolumn{3}{c|}{Model}                      & \multicolumn{2}{c|}{$k_u=0$}              & \multicolumn{2}{c|}{$k_u=0.2$}   & \multicolumn{2}{c|}{$k_u=1$}                           & \multicolumn{2}{c|}{$k_u=0$}                       & \multicolumn{2}{c|}{$k_u=0.2$}  &
\multicolumn{2}{c}{$k_u=1$}                \\ 

\multicolumn{3}{c|}{}                  & RMSE        & \multicolumn{1}{c|}{MAE} & RMSE                 & \multicolumn{1}{c|}{MAE}        & RMSE                 & \multicolumn{1}{c|}{MAE} & RMSE      & \multicolumn{1}{c|}{MAE}      & RMSE     & \multicolumn{1}{c|}{MAE} & RMSE         
& \multicolumn{1}{c}{MAE}         \\ \midrule
\multicolumn{3}{c|}{ICODE (Ours)}          & 0.048       & 0.18                    & \textbf{0.06}               & \multicolumn{1}{c|}{\textbf{0.23}} & \textbf{0.081} &  \textbf{0.30}   &  0.31                    & \multicolumn{1}{c|}{0.67}                         & \textbf{0.26}          & \multicolumn{1}{c|}{\textbf{0.82}}  & \textbf{0.50} &   \textbf{1.17}      \\
\multicolumn{3}{c|}{CDE}             & 0.048     & 0.18                    & 0.10                & \multicolumn{1}{c|}{0.36}  &  0.41 & 1.49   & 0.27                     &  \multicolumn{1}{c|}{0.60}                      & 0.30          & \multicolumn{1}{c|}{0.91}    & 0.80 &  2.70          \\
\multicolumn{3}{c|}{NODE}            & 0.049       & 0.18                    &  0.073              & \multicolumn{1}{c|}{0.28}  & 0.28 &  1.04  & 0.32                    & \multicolumn{1}{c|}{0.71}                      & 0.29         & \multicolumn{1}{c|}{ 0.92}   & 0.81 &     2.70         \\
\multicolumn{3}{c|}{ANODE}               & 0.059       & 0.20             & 0.082              & \multicolumn{1}{c|}{0.30} &  0.28 & 1.03 & 0.39                    &  \multicolumn{1}{c|}{0.82}                         & 0.30        & \multicolumn{1}{c|}{0.89}  & 1.03 &    2.83       \\

\bottomrule
\end{tabular}}
\label{tab:comparek_u}
\end{table*}

\subsection{Piecewise Inputs}
\begin{table}
\small
\caption{Performance comparison between CDE and ICODE under various piecewise inputs.}
\centering
\renewcommand{\arraystretch}{1.2}
\setlength{\tabcolsep}{1.3mm}{
\begin{tabular}{ccccc|cccc|cccc|ccc}

\cmidrule[0.8pt]{1-9} & \multicolumn{2}{c|}{}  
 & \multicolumn{6}{c}{Single-link Robot}            \\ \cmidrule{4-9} 
\multicolumn{3}{c|}{Model}                      & \multicolumn{2}{c|}{$\Delta t =0.1$}              & \multicolumn{2}{c|}{$\Delta t =0.05$}   & \multicolumn{2}{c}{$\Delta t=0.01$}                                   \\ 
\multicolumn{3}{c|}{}                  & RMSE        & \multicolumn{1}{c|}{MAE} & RMSE                 & \multicolumn{1}{c|}{MAE}        & RMSE                 & \multicolumn{1}{c}{MAE}     \\ \midrule
\multicolumn{3}{c|}{ICODE (Ours)}          & \textbf{0.113}       & \textbf{0.407}                    & \textbf{0.116}               & \multicolumn{1}{c|}{\textbf{0.490}} & \textbf{0.108} &  \multicolumn{1}{c}{\textbf{0.437}}                            \\
\multicolumn{3}{c|}{CDE}             & 0.182     & 0.683                    & 0.395               & \multicolumn{1}{c|}{1.359}  &  0.511 & \multicolumn{1}{c}{1.556}                             \\
\bottomrule
\end{tabular}}
\label{tab:compare_various_shape_u}
\end{table}

Both ICODEs and CDEs possess the capability to integrate incoming input information. However, our method demonstrates superior performance, as shown in the previous tasks. In this section, we perform a detailed investigation of their differences by examining three types of piecewise input on the task of the single-link robot. These scenarios involve inputs $u_1, u_2, u_3$ with varying shapes, as illustrated in Fig. \ref{Fig: differentU}. The values of these inputs range from $0$ to $1$. The differences among them are characterized by the length of switching, denoted by $\Delta t$. Specifically, $u_1$, the smoothest input, transitions from $0$ to $1$ over $0.1s$; $u_2$ and $u_3$ transition over $0.5s$ and $0.01s$, respectively. It is evident that the input curves become steeper as $\Delta t$ decreases. The learning results are presented in Table \ref{tab:compare_various_shape_u}. 

In the first scenario, with input $u_1$, there is no significant difference between the performance of ICODE and CDE. In the second scenario, where $u_2$ exhibits steeper transitions, the performance gap between the two methods widens considerably, with the prediction error of ICODE being approximately one-third of that of CDE. In the third scenario, where $u_3$ has the sharpest slopes during certain intervals, ICODE significantly outperforms CDE, with the prediction error of ICODE being approximately 20\% of that of CDE. It is clear that as $\Delta t$ decreases, CDE's performance progressively deteriorates, whereas ICODE remains unaffected. Notably, as $\Delta t$ approaches zero, corresponding to a "periodic step function" case, our method ICODE exhibits substantial advantages over CDE in the learning and prediction of physical models. Similarly, when external inputs exhibit rapid changes or possess infinite derivatives at certain points, CDEs tend to present a significant decline in performance or may even lose efficacy.

\begin{figure*}
\centering	
\includegraphics[width=1\textwidth]{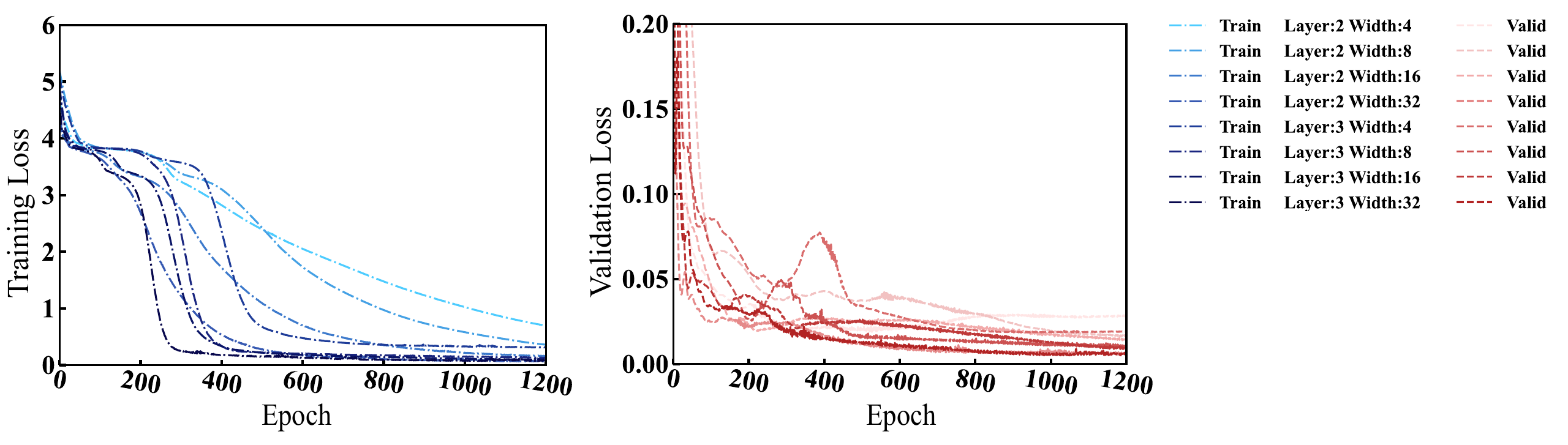} 
	\caption{The loss curves for different hidden layer numbers and widths, reflect how changes in configuration impact model performance.}
\label{Fig: fig15}
\end{figure*}

\section{Model Scaling Experiments}

ICODE has demonstrated significant superiority over other NODE-based models in the experiments conducted. In the scaling experiments, we focus on examining the scalability and architectural robustness of ICODE in comparison to the NODE model. Using the R-F equation as the task, as detailed in Section \ref{experiment}, we assess the impact of varying the width and number of hidden layers on system performance. For each scenario, we conducted $1200$ epochs on $10$ trajectories over a $1s$ duration, utilizing $75$ time points for training and $25$ for prediction. We recorded the average corresponding training loss and prediction error on each trajectory to evaluate performance.

The scatter plot in Fig. \ref{Fig: scatter} presents the training loss and prediction error for different numbers of hidden layers in the neural networks $f_i$ ($1$, $2$, and $3$), with each layer’s width fixed at 60. The results demonstrate a strong correlation between training loss and prediction error in ICODE. Notably, across all configurations, ICODE consistently achieves lower training loss and prediction error compared to the NODE model. The heatmap in Fig. \ref{Fig: heatmap} further reveals that increasing the width and number of hidden layers in ICODE enhances both training and prediction performance.

To further investigate the performance of ICODE in a more complex environment, the Glycolytic-glycogenolytic pathway model is employed as the test scenario. Experiments are conducted using various network architectures, including $2-$ and $3-$layer configurations, with network widths ranging from $4$ to $32$. The results are presented in Fig. \ref{Fig: fig15}. 
Blue lines on the left represent the training loss, and red lines on the right represent the validation loss.
For a fixed number of layers, darker lines correspond to more complex network structures. 
In the two subplots, generally, the deeper the color of the line, the lower the loss. This demonstrates that increasing the network scale effectively reduces training loss and achieves lower validation loss, resulting in more accurate predictions. These findings highlight a scaling law in ICODE, suggesting that problems of varying complexity can be addressed by appropriately adjusting the network structure.

\section{Conclusion}
In this work, we introduced a new class of neural ODEs: \emph{Input Concomitant Neural ODEs (ICODEs)}, which feature a novel structure equipped with contraction properties, incorporating the coupling of state and external input while maintaining high scalability and flexibility. The experimental results highlight the superiority of ICODEs in modeling nonlinear dynamics influenced by external inputs, with a marked advantage in scenarios involving atypical inputs. This effectiveness is evident even in systems governed by partial differential equations. 

\textbf{Limitations and Future Work}: The effectiveness of the inductive biases in ICODEs is context-dependent and may not always be preferable. Specifically, in systems where the state evolution is governed by independent input (\emph{i.e.}, no coupling between state and input), the performance of ICODEs may degrade to the level of standard neural ODEs (NODEs).  We believe this work opens promising avenues for future research. To expand the applicability of ICODEs to physical dynamics and better capture the frequent mutual interactions between state and input, we can consider more general NODEs of the form $\dot{x} = f(x,u)$, with a guaranteed contraction property. This approach would apply to a broader range of dynamics and could enhance the accuracy of learning and predicting systems with more complex characteristics and structures. Additionally, ICODEs may be more sensitive to changes in learning rates due to their potentially complex architectures. Future research will investigate this sensitivity and explore adaptive methods for learning rate adjustment and model simplification.




\bibliographystyle{IEEEtran}
\bibliography{ICODE}

\hfill

\newpage
\section*{Appendix}
\addcontentsline{toc}{section}{Appendix}

\subsection*{A. A brief introduction of existing neural ODEs and their comparison with ICODE}
\addcontentsline{toc}{subsection}{A. A brief introduction of existing neural ODEs and their comparison with ICODE}
In this section, we briefly introduce the NODE, CDE, and ANODE and clarify the core distinctions between CDE and ICODE.

\subsubsection*{1) Neural Ordinary Differential Equations}
The Neural Ordinary Differential Equation (NODE or neural ODE) is defined as follows \cite{chen2018neural}:  
\begin{equation}\label{node1_ae}  
y(0) = y_0, \quad \frac{\mathrm{d}y}{\mathrm{d}t}(t) = f_\theta(t, y(t)),  
\end{equation}  
where \( y_0 \in \mathbb{R}^{d_y} \) denotes the initial condition, \( \theta \) is a vector of learnable parameters, and \( f_\theta : \mathbb{R} \times \mathbb{R}^{d_y} \rightarrow \mathbb{R}^{d_y} \) represents a neural network. In many cases, \( f_\theta \) is implemented using a standard neural network architecture, such as a feedforward or convolutional network.

\subsubsection*{2) Neural Controlled Differential Equation}
The Neural Controlled Differential Equation (CDE) enhances classical NODEs from a mathematical perspective, enabling the effective processing of incoming data.  

The state of a neural ODE can be expressed in the form of the following Riemann integral:  
\begin{equation}\label{node2_ae}  
y(0) = y_0, \quad y(t) = y(0) + \int_0^t f_\theta(s, y(s)) \, \mathrm{d}s,  
\end{equation}  
which naturally suggests an extension to a Riemann-Stieltjes integral:  
\begin{equation}\label{cde_ae}  
y(0) = y_0, \quad y(t) = y(0) + \int_0^t f(y(s)) \, \mathrm{d}x(s),  
\end{equation}  
where \( x \) is a high-dimensional function with respect to time \( t \).  

A key limitation of ODEs, as given in \eqref{node1_ae} and \eqref{node2_ae}, is that their solutions are solely determined by the initial condition \( y(0) \) and the function $f_{\theta}$, with no direct mechanism for incorporating data that arrives at later time steps. In contrast, in \eqref{cde_ae}, the integrator \( x \) is explicitly time-dependent, allowing temporal information to be naturally incorporated into the system. This fundamental distinction underpins the development of CDEs.  

We now introduce the formal definition of CDEs: 

Let \( T > 0 \) and let \( d_x, d_y \in \mathbb{N} \). Let \( x \colon [0,T] \to \mathbb{R}^{d_x} \) be a continuous function of bounded variation. Let \( f \colon \mathbb{R}^{d_y} \to \mathbb{R}^{d_y \times d_x} \) be Lipschitz continuous. Let \( y_0 \in \mathbb{R}^{d_y} \). A continuous path \( y \colon [0,T] \to \mathbb{R}^{d_y} \) is said to solve a controlled differential equation, controlled or driven by \( x \), if it satisfies Equation \eqref{cde_ae}
for \( t \in (0,T] \).
Here, \( \mathrm{d}x(s) \) denotes a Riemann-Stieltjes integral, and \( f(y(s)) \, \mathrm{d}x(s) \) refers to a matrix-vector multiplication. 

As can be seen, CDE incorporates incoming data by integrating the known information $x(s)$. Given the differentiable assumption, a CDE is solved by reducing it to an ODE \cite{kidger2020neural,kidger2022neural}. To be more specific, let
\begin{equation*}
g_{\theta,x}(y,s) = f_\theta(y) \frac{\mathrm{d}x}{\mathrm{d}s}(s),
\end{equation*}
so that for \( t \in (0,T] \),
\begin{equation*}
\begin{aligned}
y(t) &= y(0) + \int_0^t f_\theta(y(s)) \, \mathrm{d}x(s) \\
&= y(0) + \int_0^t f_\theta(y(s)) \frac{\mathrm{d}x}{\mathrm{d}s}(s) \, \mathrm{d}s \\
&= y(0) + \int_0^t g_{\theta,x}(y(s), s) \, \mathrm{d}s.
\end{aligned}
\end{equation*}

Suppose that we have known information \(x\) with $x(t_i) := (t_i,(u_i)^\top)^\top$. Then we have \( g_{\theta,x}(y,s) = f_\theta(y) \frac{\mathrm{d}x}{\mathrm{d}s}(s) = [f_{\theta,1}(y)  \; f_{\theta,2:}(y)] [1, (\frac{\text{d} u}{\text{d} t})^\top]^\top(s) =f_{\theta,1}(y) +f_{\theta,2:}(y) \frac{\text{d} u}{\text{d} t}(s) \), where $f_{\theta,1}$ and $f_{\theta,2:}$ are neural networks to be trained. This gives
\begin{equation*}
     \frac{\text{d} y}{\text{d} t} = f_{\theta,1}(y) +f_{\theta,2:}(y) \frac{\text{d} u}{\text{d} t}(t) \quad \text{ (CDE) }.
\end{equation*}

Compare it with our ICODE:
\begin{equation*}
     \frac{\text{d} y}{\text{d} t} = f(y) +g(y) u(t) \quad \text{ (ICODE) },
\end{equation*}
where $f$ and $g$ are neural networks to be trained.
It can be seen that the difference lies in whether the derivative operation is necessary before incorporating the external information $u$ into the model.

\subsubsection*{3) Augmented Neural ODE}
The Augmented Neural Ordinary Differential Equation (ANODE) has been shown to learn representations that preserve the topology of the input space, resulting in certain functions that the NODE cannot represent. To address this limitation, ANODE \cite{dupont2019augmented} augments the space in which the Ordinary Differential Equation (ODE) is solved, enabling the model to leverage the additional dimensions for learning more complex functions. Specifically, the NODE \eqref{node1_ae} is enhanced to the following form:
\[
h(0) =
\begin{bmatrix}
y(0) \\
\mathbf{0}
\end{bmatrix} \in \mathbb{R}^{d_y+d_a}, 
\quad
\frac{\mathrm{d}}{\mathrm{d}t} h(t) = \tilde{f}_{\theta} \left( h(t), t \right).
\]
Here, \(d_a\) represents the augmented dimensions. 

ANODE is especially advantageous when the continuous system includes latent variables or unmodeled dynamics, which are common in dynamical systems with non-unique solutions. However, in the absence of such complexities, ANODE may result in diminished generalization performance due to its excessive flexibility. When the order and state variables of a continuous system are well-defined, incorporating topology preservation as prior information may be disrupted by ANODE. Unlike classification tasks that utilize dynamic systems, our focus is on modeling dynamic systems directly. Consequently, augmenting additional variables is often unnecessary in this context. As demonstrated in the experimental section, while ANODE may exhibit minimal error on the training set, its performance tends to be suboptimal on the test set.
significantly weaker.

\begin{table*}[t]
\small
\caption{Hyperparameter settings on the seven experimental tasks.}
\label{tab: Hyperparameter}
\centering
\renewcommand{\arraystretch}{1.2}

\setlength{\tabcolsep}{2.1mm}{
\begin{tabular}{ccccccccc|cccc}
\toprule
\cmidrule[0.8pt]{1-9} 
\multicolumn{3}{c|}{Task}
&   Learning Rate   & Epochs   & Network Width &   Hidden  Layers   & Trajectories   & \multicolumn{1}{c}{Time Interval}             \\ 
\midrule
\multicolumn{3}{c|}{Single-link Robot}          & $5\times 10^{-3}$       & 100                   & 50              & \multicolumn{1}{c}{3} & 10 &  \multicolumn{1}{c}{$[0,1]$}                 \\
\multicolumn{3}{c|}{DC-DC Converter}             & $5\times 10^{-4}$     & 600                   & 60               & \multicolumn{1}{c}{3}  &  10 & \multicolumn{1}{c}{$[0,1]$}                      \\
\multicolumn{3}{c|}{Rigid Body}             &  $1\times 10^{-3}$     & 200                   & 60          & \multicolumn{1}{c}{2}  &  10 & \multicolumn{1}{c}{$[0,1]$}                    \\
\multicolumn{3}{c|}{R-F Equation}             & $5\times 10^{-4}$   & 800                    & 60             & \multicolumn{1}{c}{3}  &  40 & \multicolumn{1}{c}{$[0,1]$}  \\          
\multicolumn{3}{c|}{Glycolytic-glycogenolytic Pathway}             & $5\times 10^{-4}$    & 500                    & 60             & \multicolumn{1}{c}{3}  &  10 & \multicolumn{1}{c}{$[0,2]$}  \\ 
\multicolumn{3}{c|}{Swing Equation (multi-agent)}             & $5\times 10^{-4}$    & 200                    & 60             & \multicolumn{1}{c}{3}  &  128 (batch size 16) & \multicolumn{1}{c}{$[0,5]$}  \\ 
\multicolumn{3}{c|}{Heat Conduction (PDE)}             & $2\times 10^{-3}$   & 400                    & 200              & \multicolumn{1}{c}{3}  &  5 & \multicolumn{1}{c}{$[0, 0.018]$}  \\ 
\bottomrule
\end{tabular}}
\end{table*}

\subsection*{B. Specific details about the experiments}
\begin{figure}[h]
\centering
\includegraphics[width=0.3\textwidth]{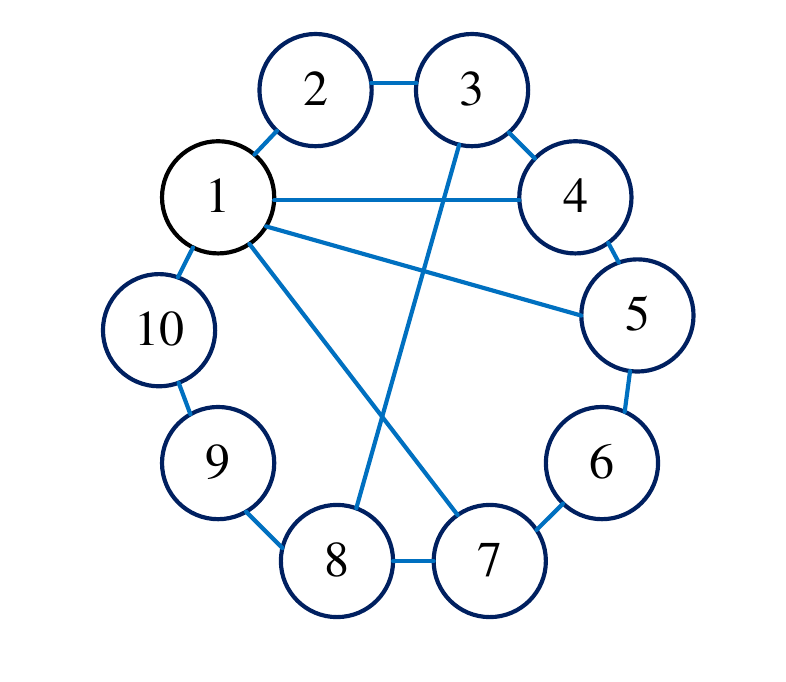}
\caption{The topology structure of the power system network.}
\label{Fig: network-topo}
 \end{figure}

\begin{figure}[h]
\centering
\includegraphics[width=0.5\textwidth]{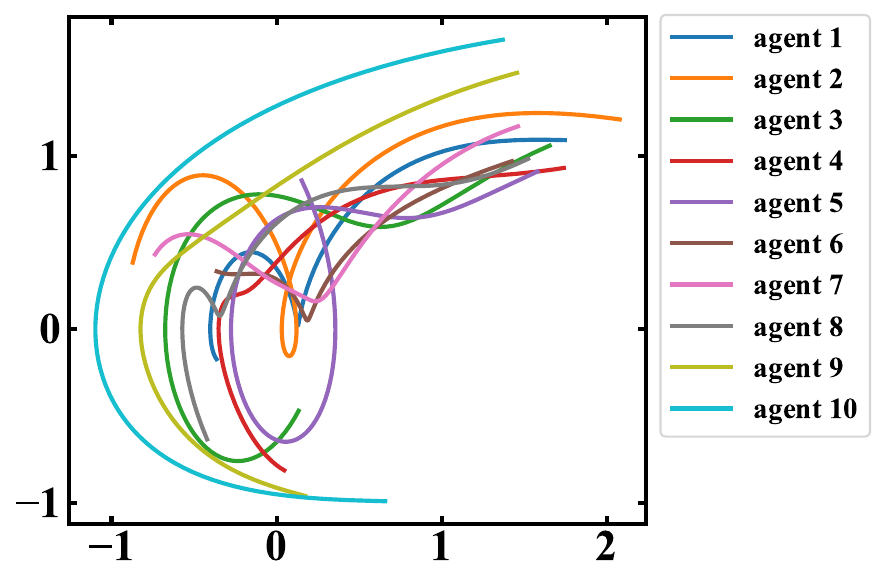}
\caption{The trajectories of the agents (synchronous generators) in the network.}
\label{Fig: network-traj}
 \end{figure}

\addcontentsline{toc}{subsection}{B. Specific details about the experiments}
The topology structure of the $10$ nodes in the experiment of the Swing Equation is shown in Fig. \ref{Fig: network-topo}. The trajectories of these agents (synchronous generators) are presented in Fig. \ref{Fig: network-traj}.

\subsection*{C. Hyperparameter settings}
\addcontentsline{toc}{subsection}{C. Hyperparameter settings}

This section provides a comprehensive description of the hyperparameters used in the experiments, as summarized in Table \ref{tab: Hyperparameter}. To ensure a fair comparison, all methods are evaluated under identical experimental conditions. Specifically, \textit{Hidden Layers} denotes the number of hidden layers in the neural networks, while \textit{Network Width} represents the number of neurons per layer. The \textit{Trajectories} column specifies the number of trajectories used for model training, and \textit{Time Interval} indicates the temporal length of these trajectories.  

For the first four experiments, the ANODE model is augmented with $2$ additional dimensions. In the fifth experiment, this augmentation is increased to $10$ dimensions. In subsequent experiments, the ANODE model is expanded to $200$ dimensions to accommodate the increased complexity of the partial differential equation (PDE) case. The initial augmented states are generated using a multi-layer perceptron (MLP) with a single hidden layer.  

The implementation employs the Softplus activation function and the Adam optimizer. Model evaluation is performed every $10$ epochs, with the model achieving the lowest mean squared error (MSE) retained for final analysis.

\newpage

\end{document}